\documentclass[]{fairmeta}

\usepackage{gacha}

\title{Gacha Decoding: Eliciting Diverse Generations Through Instruction Following}
\author[1,2]{Scott Geng}
\author[1]{Yufei Zhang}
\author[1]{Joseph Lee}
\author[1,*]{Jerry Li}
\author[2,*]{Marjan Ghazvininejad}
\author[1,*]{Pang Wei Koh}

\affiliation[1]{University of Washington}
\affiliation[2]{Meta Superintelligence Labs}

\contribution[*]{Joint last authors}

\abstract{
We introduce \ours, an inference-time method for eliciting diverse language model generations that scales with model capability. Across open-ended domains (in-the-wild chat, creative writing, planning for image generation, and protein design), \ours significantly outperforms existing generation diversity approaches at equal quality (up to 2.4x Vendi over the next-best prior approach), reaching the same number of high-quality modes with over an order of magnitude fewer samples (11.0x) and discovering novel modes that no other approach surfaces. 
Our key insight is to treat diversity as an instruction-following problem: rather than relying on the LM's token entropy, we combine its instruction-following capability with randomness from an external RNG tool to scalably identify and realize distinct modes of the response space. 
This approach of ``planning with dice'' enables Gacha to invert the long-observed tension between diversity and model capability.
As the underlying LM becomes a better instruction follower, diversity under \ours consistently improves---even as its token entropy and diversity under prior approaches decline. Together, our results highlight that instruction following, rather than token entropy alone, can drive generation diversity.
}
\date{\today}
\correspondence{Scott Geng at \email{sgeng@cs.washington.edu}}

\begin{document}

\maketitle
\begin{figure}[b!]
    \centering
    \includegraphics[width=0.96\linewidth]{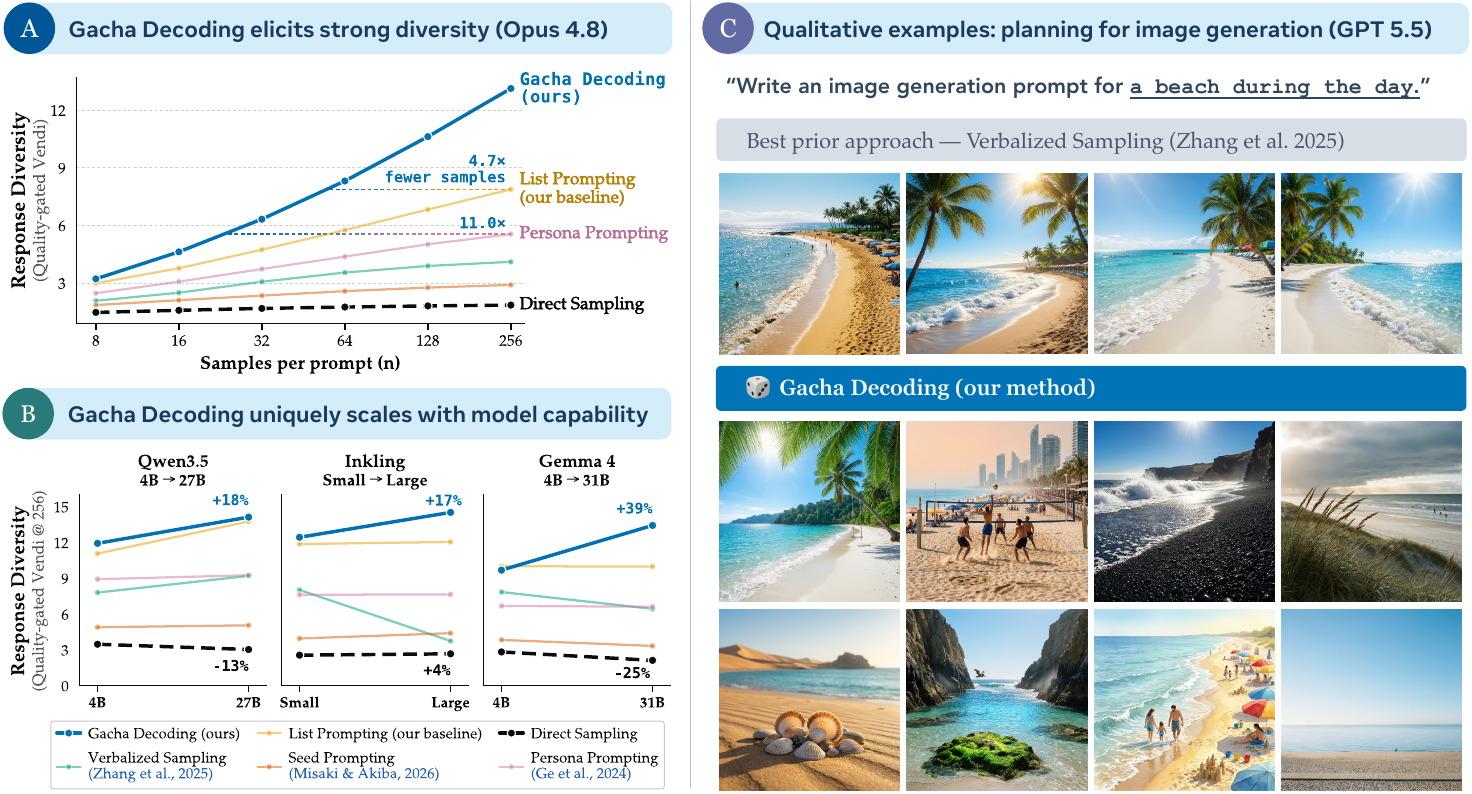}
    \caption{
    We introduce \textbf{\ours}, an inference-time method that strongly improves LM response diversity (\textbf{A}). We recast diversity as an instruction-following problem: instead of relying on token entropy, Gacha pairs a model's IF capabilities with external RNG to identify and realize distinct output modes. This reformulation enables Gacha to improve with stronger models (\textbf{B}), even when diversity under prior approaches stagnates or declines. (\textbf{A--B}): results on InfinityChat~\citep{jiang2026artificial}. (\textbf{C}): random sampled images from text prompts generated by ours and baselines.
    }
    \label{fig:teaser}
\end{figure}

\section{Introduction}
\label{sec:intro}
Language model (LM) systems are increasingly entrusted with creative or exploratory tasks (writing, ideation, data generation, and scientific discovery, to name a few), but they often lack the human-like diversity and novelty these applications demand. Modern models collapse onto just a few typical responses, even on open-ended prompts where we expect many possibilities~\citep{jiang2026artificial}. Left unchecked, this shortfall lowers user utility~\citep{zhang2025noveltybench}, constrains practitioners who rely on diverse samples to improve synthetic data or scale test-time compute~\citep{cui2025entropy}, and in the long run may risk homogenizing our thinking~\citep{abdulhai2026llms,yakura2024empirical}.

Spurred by this gap, researchers have made a flurry of attempts at improving generation diversity, often treating diversity as a problem of limited entropy in the LM's next-token distribution. Efforts have hence focused on broadening or reshaping this distribution, for example, by changing sampling parameters~\citep{ahmed2025intent}, applying entropy-encouraging losses~\citep{shi2018toward,li2025jointly,sorensen2026spectrum}, or curating diverse training data~\citep{zhang2024forcing,springer2026annotations}. While these approaches have shown gains, this distribution-centric paradigm of diversity has led to an apparent tension between response diversity and model capabilities, as capabilities are often improved via post-training methods that reduce entropy as a side effect~\citep{cui2025entropy}. As models get better, diversity appears to be getting worse~\citep{zhang2025noveltybench}.

We argue that this tension is not fundamental. In this paper, we recast generation diversity as an instruction-following (IF) problem: we show that instruction following, as opposed to token entropy, can serve as the engine driving response diversity. In particular, we show how to harness a model's IF capabilities together with external randomness to scalably identify and realize distinct modes of the response space. This view crystallizes an emerging thread in the literature; many recent methods implicitly leverage instruction following for diversity, for example by prompting to steer generation toward less typical responses~\citep{zhang2026you,misaki2026string,wu2025mode,zhang2025verbalized}. By treating diversity explicitly as an IF problem, we fundamentally shift the LM's role from maintaining diffuse distributions to faithfully executing instructions---a capability that can improve with better alignment, even if the token distribution sharpens. 

We instantiate this view as \textbf{\ours}\footnote{Named after \textit{Gachapon} vending machines that dispense random toys from a large curated pool.}, our inference-time method for eliciting diverse LM generations. Across open-ended chat, creative writing, planning for image generation, and protein design, \textbf{\ours substantially outperforms prior best methods}, yielding up to a 2.4x gain in Vendi-scored diversity at similar or higher quality (\hyperref[fig:teaser]{\cref*{fig:teaser}A}). Most strikingly, \ours's diversity consistently scales with model capability, even when diversity with direct sampling or other prior methods declines (\hyperref[fig:teaser]{\cref*{fig:teaser}B}). Gacha thus inverts the typical diversity-capability tradeoff, reflecting its reduced reliance on token entropy. As an extreme illustration of this, Gacha's diversity gains persist even under deterministic greedy decoding (\cref{sec:mainexps:if_greedy}).

\begin{figure}[t]
    \centering
    \includegraphics[width=0.8\linewidth]{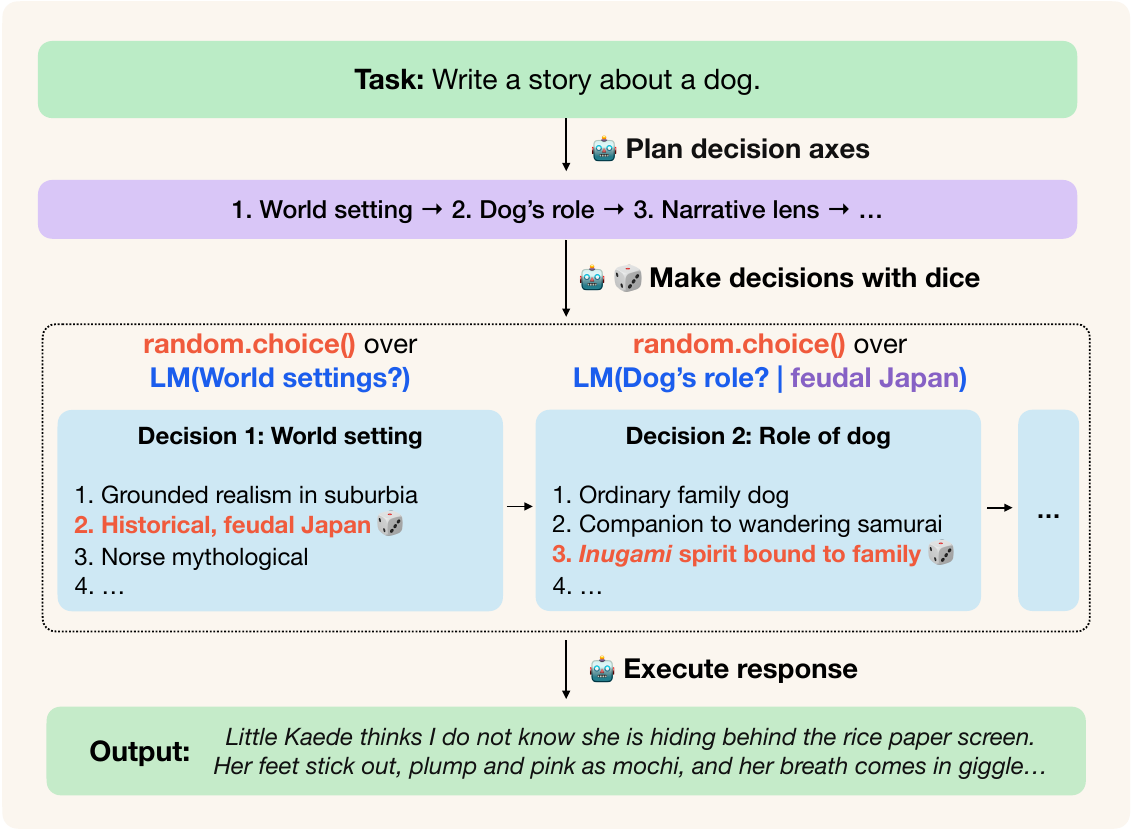}
    \caption{\ours generates responses by ``planning with dice.'' Given a task, we instruct the LM to identify and make a sequence of decisions to determine the final response. At each step, the LM enumerates possible options, and uses external randomness to select one uniformly. This design utilizes the LM's IF capabilities while minimizing reliance on token entropy.}
    \label{fig:methods}
\end{figure}

How does \ours work? At a high level, we ``plan with dice'' (\cref{alg:gacha,fig:methods}). Given a task, we instruct the LM to identify a sequence of semantic decisions that shape the output (e.g., to write a story, the setting, characters, and so on). We then sample these decisions sequentially using external RNG to pick from LM-enumerated options, and realize the resulting randomized plan into a final response. Crucially, we condition each decision on those already sampled so that each random choice reshapes the downstream options, allowing variation to compound combinatorially without token stochasticity. Taken together, we view \ours as autoregressive generation lifted from tokens to ideas: each step samples a semantic choice rather than a token, conditioned on all prior choices. In practice, Gacha can be implemented as a lightweight harness around any existing LM.

Overall, our work engages an ongoing dialogue between (a) viewing LMs as classical generative models directly responsible for maintaining a diverse distribution, and (b) viewing them as components of larger AI systems. In a system, variation need not be encoded in the LM's distribution: we can source randomness externally and harness the model's IF capability to propagate it into useful, diverse responses. \textbf{Our conceptual contribution} is to crystallize the view that generation diversity can be treated as an instruction-following problem. \textbf{Our technical contribution} is to show how planning with dice operationalizes this principle, yielding strong empirical results. Our inference-time results also suggest new training directions, including end-to-end optimization with our harness or distillation of our diverse generations back into the underlying model. We now detail \ours (\cref{sec:method}), compare it against prior methods (\cref{sec:mainexps}), and analyze why it works (\cref{sec:analysis}).
\section{\ours: Diversity Through Instruction Following}
\label{sec:method}

\vspace{-2mm}
\subsection{Overview: What is \ours?}
We present \ours, our method for eliciting diverse generations from language models. We first describe how \ours works, and then explain the three key design elements underlying its success. We show the core algorithm in \cref{alg:gacha} and a concrete example in \cref{fig:methods}. To summarize, the overall idea is to augment an LM's reasoning with randomness. Given a task, the LM identifies a sequence of decisions that shape the output, then samples each decision by (1) enumerating plausible options and (2) selecting one with external RNG. For example, to write a story about a dog, we might first decide the setting, using the LM to list distinct options (e.g. modern suburbia, historical Japan, etc.) and picking one with dice. Later decisions---such as the dog's role or narrative perspective---are conditioned on that sampled choice, allowing diversity to compound across the decision trajectory. By using instruction following to enumerate options and external randomness to select among them, Gacha shifts the source of diversity away from token entropy. Overall, as we will detail more below, we can view this procedure as a randomized autoregressive process in idea-space. In practice, \ours can be implemented as a simple agentic harness over an existing LM; see~\cref{appdx:method} for one concrete instantiation.

\newcommand{\Phase}[1]{\Statex\hspace*{-\leftmargin}\textbf{#1}}

\begin{algorithm}[htbp]
\caption{\ours: diversity through instruction following by planning with dice}
\label{alg:gacha}
\begin{algorithmic}[1]
\Require Prompt $x$; language model $\mathrm{LM}$; decision budget $T$; external RNG
\Ensure A diverse response $y$
\vspace{1mm}

\Phase{Phase 1: Plan the decision sequence.}
\State Instruct the LM to identify $T$ semantic decision axes $(a_1,\ldots,a_T)$ that will shape the response. At this stage the model specifies only \textit{which} decisions will be made, not their values.
\vspace{1mm}

\Phase{Phase 2: Autoregressively sample the decisions.}
\For{$t = 1, \ldots, T$}
  \State \textbf{Enumerate options.} Instruct the LM to list a set $\mathcal{C}_t$ of plausible, semantically distinct options for decision axis $a_t$, conditioned on the prompt and all decisions committed so far:
  \Statex \hfill $\mathcal{C}_t \leftarrow \mathrm{LM}(x,\, a_t,\, z_{<t})$ \hfill\null
  \State \textbf{Roll dice.} Sample one option uniformly with external RNG and commit it to the trajectory:
  \Statex \hfill $z_t \sim \mathrm{Uniform}(\mathcal{C}_t)$ \hfill\null
\EndFor
\vspace{1mm}

\Phase{Phase 3: Realize the response.}
\State \textbf{Integrate.} Combine all decisions $z_{1:T}$ into a final plan: instruct the LM to propose several full plans, and select one uniformly with external RNG.
\State \textbf{Generate.} Instruct the LM to write the final response $y$ from the prompt, the decision trajectory, and the selected plan, so that $y$ is consistent with every sampled choice:
\Statex \hfill $y \leftarrow \mathrm{LM}(x,\, z_{1:T},\, \text{plan})$ \hfill\null
\State \Return final response $y$
\end{algorithmic}
\end{algorithm}

\subsection{Diversity From Instruction Following by Planning with Dice}
We explain the three design choices that shift \ours's source of diversity from token entropy to instruction following: (1) work in semantic idea space, (2) sample ideas by decoupling enumeration from selection, and (3) sample ideas autoregressively. As we will later show in our ablations (\cref{sec:analysis}), each component is strictly necessary. Now, to first motivate these designs conceptually, let us build them up one at a time, starting from the simplest baseline: direct sampling from an LM. Suppose we are given a prompt $x$ and generate a response $y$ using standard token autoregression from an LM $p_\theta$. Here, $y$ is built as a sequence of stochastic token decisions:
\begin{equation}
    p(y\mid x) = \prod_{i=1}^T p_\theta(y_i \mid x, y_{<i}).
\end{equation}

\paragraph{Working in idea space.} Standard token sampling has two limitations for diverse generation. First, diversity fully relies on entropy in the LM's token distribution, which often becomes increasingly concentrated as models improve. Second, token variation is only a proxy for the semantic variation we care about: different token sequences may express the same idea, while unique approaches may pivot at only a few key tokens. A natural first step is then to generate in two stages, re-factorizing generation to separate a response's high-level semantics from its surface realization. Formally, we introduce a latent variable $z$, denoting a high-level idea or plan for the response, and write
\begin{equation}
p(y\mid x)=\int p_\theta(y\mid z,x)\,q(z\mid x)\,dz.
\end{equation}
Under this view, diverse generation reduces to sampling diverse semantic plans $z$, which the LM then realizes as responses. Prior methods often instantiate this idea by prompting an LM to generate the plan $z$ directly~\citep{yao2019plan,ahmed2025intent,springer2026annotations}, for example,
\begin{equation}
q(z \mid x) \approx p_\theta\left(z \mid \texttt{Give a plan for}\,\, x\right).
\end{equation}
This moves variation to the semantic level, but does not yet change its source: diversity over plans again hinges on the LM's (limited) token entropy. We therefore seek a different way to sample $z$.

\paragraph{Decoupling idea enumeration from selection.} Rather than using the LM to stochastically generate an idea directly, we use it to \textit{enumerate} a set $\mathcal{C}$ of plausible, distinct choices and delegate \textit{selection} to an external RNG, for example $z \sim \mathrm{Uniform}(\mathcal{C})$. This distinction is critical. While direct sampling relies on the LM assigning sufficient probability mass across diverse ideas, enumerating distinct ideas in-context is largely an instruction-following capability. Even under greedy decoding, a capable model can condition on what it has already generated and produce something new. We can view this random enumeration-selection process as sampling from $q(z\mid x)$ at very high effective temperature, where the support is verbalized by the LM. Variation over $z$ then depends only on which possibilities the LM can enumerate, which can improve with IF capability. Working in idea space makes this disentanglement especially natural: we can randomize aggressively over semantic choices, such as a story setting or solution strategy, without sacrificing local token coherence.

\paragraph{Sampling ideas autoregressively.}
One could stop here and simply instruct the model to list complete plans $z$ and randomly select among them. However, our diversity would be bottlenecked by the model's effective context window---when generating any list, the model can only output so many distinct options before degrading into low-quality or redundant options. Naively generating many lists with independent LM calls (e.g., \citet{zhang2025verbalized} generate parallel lists of full responses $y$) does not solve this problem because variation across lists would again depend on token stochasticity. To scale beyond this limitation, we make the idea sampling autoregressive. We factorize the distribution over ideas $q(z\mid x)$ into a sequence of conditional semantic choices:
\begin{equation}
q(z\mid x) = q(z_{1:T} \mid x) = \prod_{t=1}^T q(z_t \mid x,z_{<t}).
\end{equation}
Now, rather than using our LM to enumerate complete plans (i.e., from the joint distribution), we enumerate and randomize the next local choice conditioned on previous ones. With this autoregressive structure, each randomized choice changes the context for future LM enumeration calls, allowing us to explore combinatorially many trajectories even with a fixed number of choices at each step.  
Further, earlier choices reveal possibilities that are hard to anticipate in an unconditioned list. For example, once a story is placed in ancient Rome, subsequent decisions might involve gladiators or senate conspiracies, setting-specific elements that might never surface when listing complete story plans from scratch. Taken together, our three design choices yield \ours, which samples diverse generations by autoregressively sampling diverse semantic plans using external randomness.
\section{Main Experiments}
\label{sec:mainexps}

\vspace{-2mm}
\subsection{Experimental Setup}
We now compare \ours against existing methods on three open-ended domains: in-the-wild chat, creative writing, and prompt rewriting for image generation. We also show a pilot study applying \ours on protein design (\cref{sec:mainexps:protein}). Complete details for our evaluation domains, baselines, and experimental setup are presented in \cref{appdx:exp_setup}. Each domain consists of a set of prompts $\left\{x_1,\ldots,x_m \right\}$. To benchmark the performance of a given inference-time diverse generation method $\mathcal{M}$ with an LM $\theta$, we generate $N$ responses for each prompt $x_i$:
\begin{equation}
    \mathcal{M}_\theta(x_i) \mapsto \left\{y_i^{(1)},\ldots,y_i^{(N)}\right\}.
\end{equation}
We measure the quality of each response, and report a \textit{quality-gated} diversity score for the full response set using domain-informed metrics. Specifically, for each prompt $x_i$, we only measure diversity among responses $\{y_i^{(j)}\}$ whose quality exceeds 95\% of the average quality achieved via vanilla direct sampling on that prompt. This approach lets us normalize by quality on a prompt- and model-specific basis and prevent low-quality samples from inflating the diversity score~\citep{shypula2025evaluating}. Finally, we average metrics across all prompts of the domain. We also vary $N$ to study how each method's diversity scales with the number of generations.

\subsubsection{Models, baselines, and prior approaches}
We compare \ours against the following baselines and prior inference-time methods for diversifying model generations. See \cref{appdx:exp_setup:baselines} for full details, including prompts.

\begin{enumerate}[itemsep=-1pt,topsep=0pt]
    \item \textbf{Direct sampling.} We sample IID responses to the original prompt using each model's default temperature. We also consider \textbf{high-temperature direct sampling}, using temperature 1.5.
    \item \textbf{List prompting.} We instruct the model to produce $k$ (a hyperparameter) distinct and varied responses in a single generation. Since earlier responses remain in context while later responses are generated, this method allows the model to explicitly avoid repeating its previous outputs.
    \item \textbf{Verbalized Sampling}~\citep{zhang2025verbalized}. We ask the model to verbalize a distribution over possible responses, producing $k$ responses together with their probabilities in a single generation.
    \item \textbf{Direct Planning (Intent-Factored Generation)}~\citep{ahmed2025intent}. We generate in two stages; we first sample a high-level idea or plan from the LM at a higher-than-typical temperature (1.2), and then produce a final response conditioned on the idea in a separate call.
    \item \textbf{Seed Prompting}~\citep{misaki2026string}. We supply the model a random ASCII string (e.g. ``\texttt{aYn67}...'') and instruct the model to generate a response informed by that string. 
    \item \textbf{Persona Prompting}~\citep{ge2024scaling,santurkar2023whose}. We instruct the LM to respond as a specific persona, which we randomly sample from a large pre-specified set. Like seed prompting, persona prompting varies the prompt to steer toward varied outputs, but uses semantically meaningful personas rather than arbitrary strings.
\end{enumerate}

\textbf{Evaluated models.} We evaluate on both open-weight (GPT-OSS, Inkling, and Qwen~3.5 families) and closed-source (Claude Opus~4.8 and GPT~5.5) language models. We always enable thinking mode and use each model's recommended sampling parameters. Note that the API models do not allow altering sampling temperature with thinking mode; we thus omit the methods which require higher temperature for those models. Please see \cref{appdx:exp_setup:models} for full inference details.

\subsubsection{Open-ended evaluation domains}
We describe our evaluation domains and prompts, and how the quality and diversity of the resulting responses are measured. See \cref{appdx:exp_setup:evaluation} for complete details.

\paragraph{In-the-wild chat.} Open-ended tasks such as advice-seeking, writing, and ideation are among the most common uses of chat systems~\citep{chatterji2025people}. We evaluate on such queries through InfinityChat~\citep{jiang2026artificial}. Following the original work, we measure diversity as response embedding distance, reporting Vendi score~\citep{friedman2022vendi}. We measure quality as LM-judged response validity.

\paragraph{Creative writing.} We construct an in-the-wild creative writing benchmark by filtering WildChat~\citep{zhao2024wildchat} for fiction prompts. Following~\citet{zhang2025verbalized}, we score story quality from 0--20 using the LM-judge rubric from~\citet{paech2023eq}. We measure diversity inspired by~\citet{springer2026annotations}: we (1) annotate stories along 8 semantic attributes (\textsc{premise}, \textsc{genre}, \textsc{setting}, \textsc{main character}, \textsc{narrator pov}, \textsc{tone}, \textsc{imagery}, \textsc{resolution}), (2) compute a per-attribute embedding Vendi score, and (3) report their macro average. This decomposition captures high-level creative variation and reduces sensitivity to low-level lexical differences.

\paragraph{Image generation.} Production text-to-image systems often use an LM to expand an initial user image prompt $x$ into a detailed caption $x'$ before image generation: $x \mapsto x' \mapsto \text{final image }I$. We evaluate applying diverse generation methods at this expansion stage to increase downstream image diversity. We sample open-ended prompts $x$ from OneIGBench~\citep{chang2026oneig} and compare methods for generating diverse expanded captions $x'$, rendering images with Qwen-Image-2512~\citep{wu2025qwenimagetechnicalreport}. We measure image quality using OneIGBench's LM-judge rubrics for alignment with the original prompt $x$, and diversity using Vendi score over DreamSim embeddings~\citep{fu2023dreamsim} of the downstream images.

\begin{figure}[!t]
    \centering
    \includegraphics[width=\linewidth]{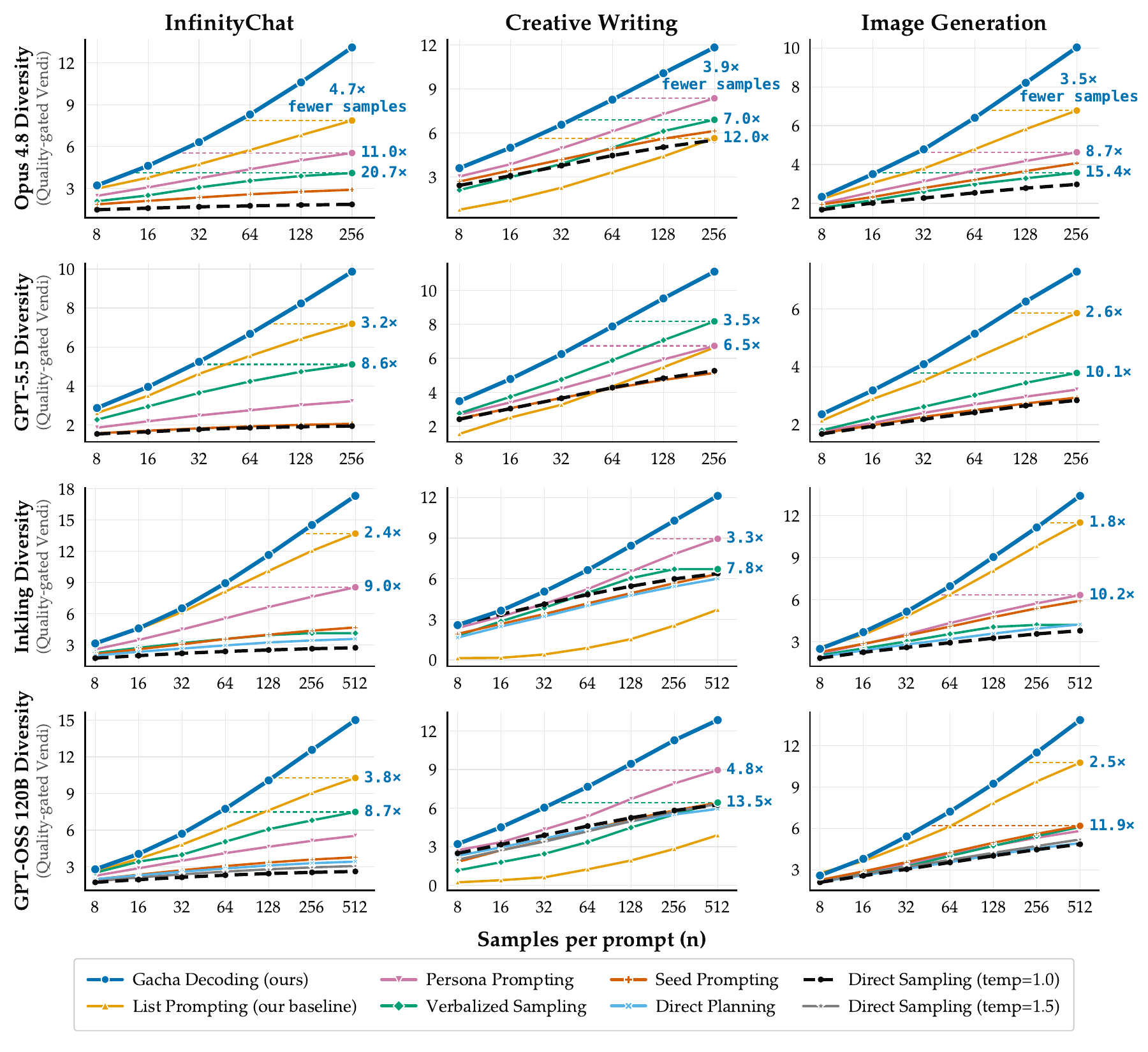}
    \vspace{-2.15em}
    \caption{\ours achieves much higher quality-controlled diversity than prior approaches across all models and task domains, with gains increasing at larger generation scales. This gain in Vendi translates to strong improvements in mode-coverage sample efficiency (dashed lines). For example, on open-ended chat with Opus 4.8, Gacha requires 11.0x fewer samples to cover the same number of modes as 256 samples from persona prompting (next-best prior approach).}
    \label{fig:mainresult}
    \vspace{-0.57em}
\end{figure}

\subsection{Main Results}
We show our main results in \cref{fig:mainresult} and highlight examples of our method's generations in \cref{fig:teaser,appdx:results:examples}. We provide extended results (including raw quality scores) in \cref{appdx:results:main}.

\paragraph{\ours strongly outperforms existing approaches, with gains continuing to increase at larger generation scales.} Our marquee result is that \ours yields significantly more diversity than prior approaches, at matched quality. For example, on Opus 4.8 with 256 generations per prompt, \ours achieves a 7.1x increase in quality-gated diversity over direct sampling on open-ended chat. To put this in a sample efficiency perspective, our method requires 4.7x fewer generations than our list prompting baseline---the next-best method---to reach the same level of effective mode coverage, matching 256 list generations with only $\sim\,$55 samples. \ours similarly achieves 11.0x gain over persona prompting, the next-best prior approach, and 20.7x and near-38x efficiency gains over Verbalized Sampling and Seed Prompting, respectively.

\paragraph{\ours can discover novel modes.} These quantitative gains often reflect the emergence of qualitatively distinct response modes that other methods never reach. For example, when prompted to ``\textit{Write a horror story about a Singapore apartment},'' Claude with \ours generates several second-person stories, while all other methods remain limited to third-person narration (\hyperref[tab:greedy_decoding]{Appendix Figure~\ref*{fig:story_qualitative}}). On image generation, the diversity gains are often very visually apparent (\cref{fig:teaser}).

\paragraph{List prompting is a strong baseline.}
Intriguingly, we find that simply prompting for a list of responses often outperforms all existing approaches we tried, in contrast with prior results~\citep{zhang2025verbalized}. We conjecture that its stronger performance in our experiments stems from (1) carefully tuning the number of generations per list as a hyperparameter, and (2) evaluating at much larger generation budgets (up to 512 responses per prompt, versus around 8–32 in prior work). 

\subsection{Gacha scales with model IF capability and works even under  greedy decoding}
\label{sec:mainexps:if_greedy}

\begin{figure}[t]
    \centering
    \includegraphics[width=\linewidth]{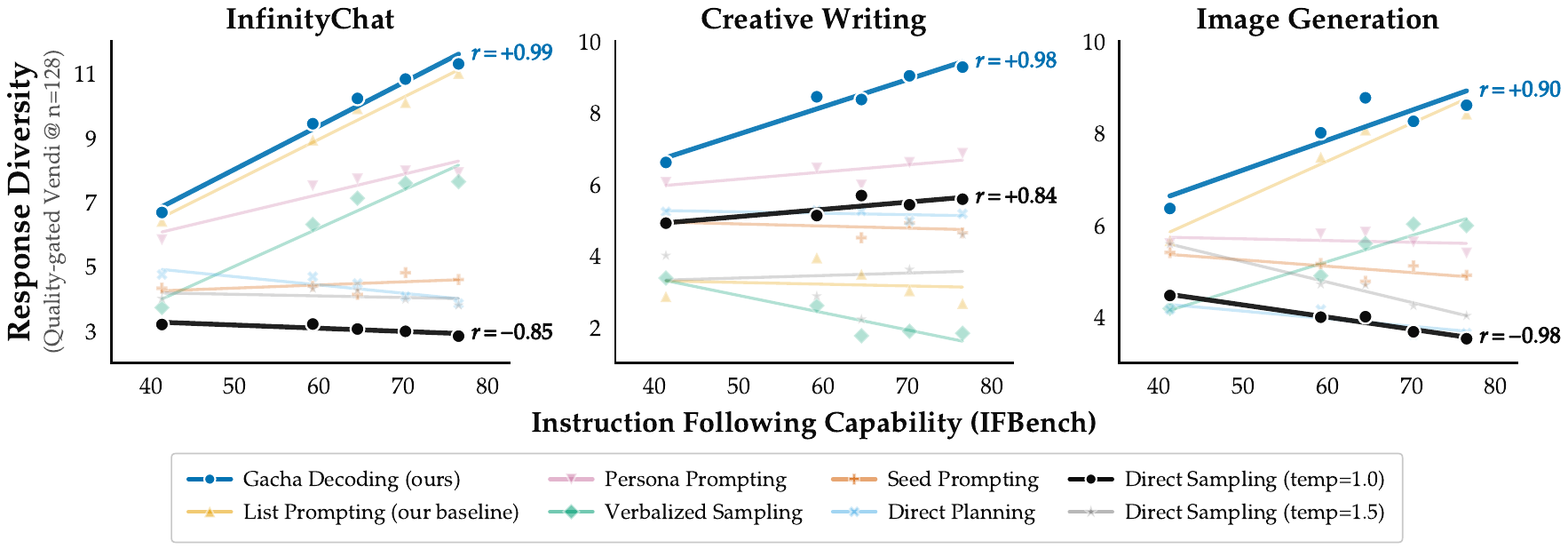}
    \caption{Our method derives diversity from an LM's instruction following ability as opposed to token entropy. Hence, our method is the only approach to consistently improve as the underlying LM's capabilities do, even when token entropy and diversity under entropy-reliant methods decline.}
    \label{fig:ifscaling}
\end{figure}


\ours is designed to derive diversity from an LM's instruction-following (IF) capability rather than its token entropy. This design suggests two predictions: (1) \ours's diversity should increase with IF capability, and (2) \ours should remain diverse even in an extreme case where the underlying LM has zero token entropy, such as under greedy decoding.

\textbf{Scaling with instruction following.} Prior work has often observed an inverse relationship between response diversity and model capability, as stronger models can have lower-entropy token distributions. In contrast, because \ours relies on instruction following, its diversity should \textit{improve} as the underlying LM improves. We test this across Qwen 3.5 models of varying parameter sizes, plotting each method's quality-gated diversity against the model's IFBench instruction following score~\citep{pyatkin2026generalizing} in \cref{fig:ifscaling} (specifically, in order of IF capability: Qwen 3.5 2B, 4B, 9B, 35B-A3B, 27B).

Overall, we find that direct-sampling diversity indeed tends to decline with larger, more capable models: on chat and image generation, diversity is strongly negatively correlated with IF capability. On writing, direct sampling exhibits a positive trend between IF and quality-gated diversity, but this gain is entirely driven by quality improving as the underlying LM improves; raw diversity is flat (5.85 $\rightarrow$ 5.79) as we go from 2B to 27B. In contrast, \ours scales significantly and positively with IF capability on both raw diversity and quality-gated diversity, and is the only method to do so consistently across all domains. Some baselines, including list and persona prompting, show mixed scaling. We view these methods as partial realizations of our IF-centric view on diversity: list prompting, for example, can utilize IF to diversify responses \textit{within} a single list, but diversity across independent lists still depends on token entropy. This works well for short-form domains like chat, where many responses can be considered within a single list, but breaks down for longer-form tasks such as creative writing. Gacha avoids this bottleneck via idea-space autoregression with external RNG, allowing IF-driven diversity to scale beyond what can be enumerated in a single call.

\textbf{Diversity under greedy decoding.} Even under greedy decoding (zero token entropy) on Qwen3.5 27B, GPT-OSS, and Inkling, we find \ours retains its strong diversity, often outperforming all baselines run \textit{with} temperature sampling. In contrast, list-based methods degrade sharply without entropy (\hyperref[tab:greedy_decoding]{Appendix Table~\ref*{tab:greedy_decoding}}).

\begin{figure}[htbp]
    \centering
    \includegraphics[width=\linewidth]{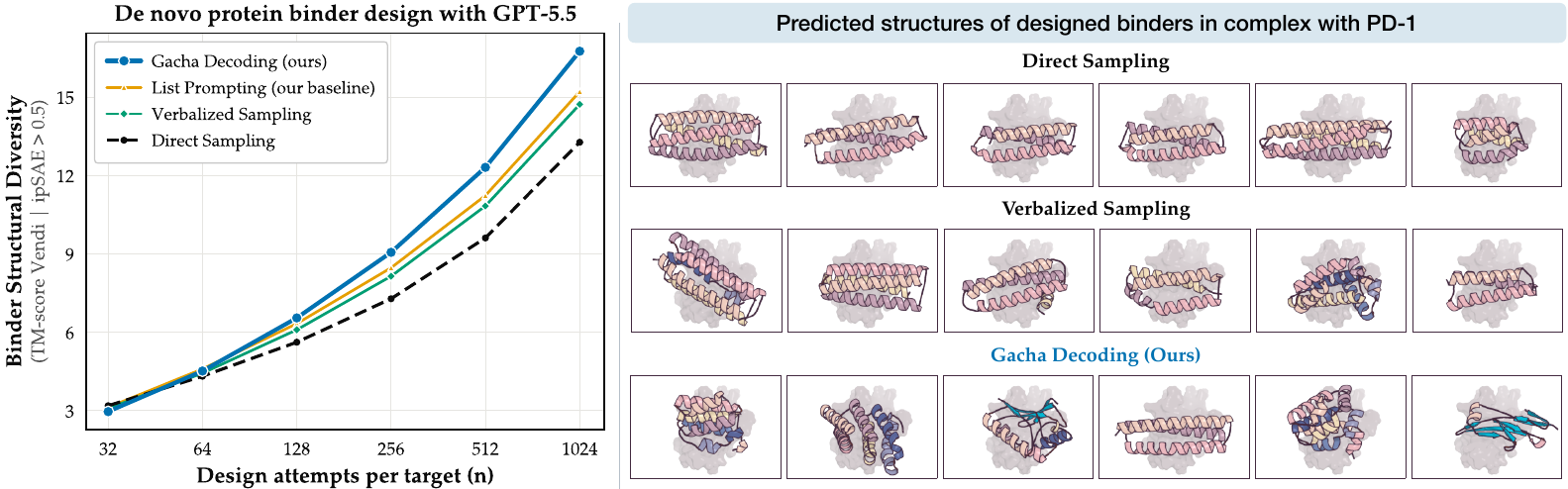}
    \caption{We use GPT 5.5 with \ours to design diverse protein binders against downstream targets. Gacha yields the most structurally-diverse binders (TM-score Vendi) among designs passing a binding-quality threshold (ipSAE$_{\min}>0.5$), with the gap widening with design budget. Against PD-1, \ours discovers binders with mixed $\alpha$/$\beta$ secondary structure ($\beta$-sheets colored cyan), while Verbalized Sampling remains limited to all-helical folds.}
    \vspace{-0.5em}
    \label{fig:protein}
\end{figure}

\subsection{A case study in protein design}
\label{sec:mainexps:protein}
As a pilot study, we apply \ours to a downstream scientific workflow that directly benefits from increased diversity: de novo protein binder design. Generative protein models are increasingly used to accelerate drug discovery by designing candidate binders against therapeutic targets~\citep{cao2022design,watson2023novo}, but their outputs often concentrate on a narrow set of structural folds~\citep{listov2024opportunities}, reducing the effective number of designs explored. Recent text-to-protein models condition on natural language protein descriptions, allowing us to improve structure diversity via caption diversity analogously to image generation. Using GPT 5.5, we generate binder descriptions with \ours and baseline methods for eight benchmark targets (e.g. PD-1, an immune checkpoint receptor targeted in cancer immunotherapy)~\citep{pacesa2025one}, and realize each description as a target--binder complex with a fixed protein-generation model~\citep{lee2026prompteina}. We refold each design with its target using ESMFold2~\citep{candido2026language}, score predicted interface confidence using ipSAE$_{\min}$~\citep{dunbrack2025res}, and retain designs with scores above 0.5~\citep{shanehsazzadeh-autonomous-protein-binder-design-2026}. Among successful designs, we quantify structural diversity using Vendi score with pairwise Foldseek TM-scores as the similarity kernel~\citep{van2024fast,zhang2004scoring}. Please see \cref{appdx:protein} for full setup details. Results in \cref{fig:protein} show that \ours yields more diverse binder structures both quantitatively and qualitatively; for example, against PD-1, Gacha produces binders with mixed $\alpha/\beta$ secondary structure, whereas Verbalized and list prompting remain restricted to all-helical folds.
\section{Analysis: Why Does Gacha Decoding Work?}
\label{sec:analysis}

We ablate our three design choices from~\cref{sec:method} to show their necessity in our method's success. Recall that Gacha autoregressively samples idea-level semantic decisions, using external RNG to randomize over LM-enumerated candidates at each local step. We consider these ablations (details in \cref{appdx:results:ablations}):
\begin{enumerate}[
    label=\textbf{(\Alph*)},
    leftmargin=*,
]
    \item Randomizing aggressively over LM-curated ideas rather than tokens. We compare Gacha to extreme-temperature token sampling, using top-$p$~\citep{holtzman2019curious} and top-$K$ decoding~\citep{fan2018hierarchical} to randomize over an analogous ``curated set'' of top tokens at each step.
    
    \item Sampling ideas by randomizing over LM-enumerated options. We consider no-RNG and no-enumeration Gacha variants. For each local decision, we instead directly prompt the model to ``make a random choice'' from its own listed options without external randomness (no-RNG), or sample an idea directly from the model via prompting (no-enumeration).

    \item Sampling ideas autoregressively. We compare against two non-autoregressive Gacha variants: (1) \textit{idea-list}, which enumerates $k$ complete ideas in a single context (i.e., samples from the joint distribution over ideas) and (2) independent axes, which samples each local decision independently, treating the joint distribution as a product of marginals rather than conditionals.
\end{enumerate}
We re-run our main experiments with these ablations on GPT-OSS-120B and Opus 4.8, finding that all variants are weaker than \ours (\cref{tab:ablations})---all our design choices are critical. To name one example, the no-RNG variant nearly halves \ours's diversity on chat: the model's own choice among its enumerated options is far from uniform and collapses toward middle-ranked options in the generated list, so true external randomness is necessary. As a further anecdote, even pretrained base models gravitate toward 37 when asked for a random number between 1 and 100, mirroring a well-known bias in humans.\footnote{Diversity is also hard for individual humans! See \url{https://www.youtube.com/watch?v=d6iQrh2TK98}, which runs an informal $n{=}200{,}000$ survey and finds humans gravitate towards 37, potentially explaining this bias in LMs.} See \cref{appdx:results:ablations} for corroborating ablation results on Inkling and GPT 5.5, plus further discussion.

\begin{table}[t]
\centering
\setlength{\tabcolsep}{6pt}
\begin{NiceTabular}{lccc}
\CodeBefore
  \rowcolor{metabg}{2,9,13}
\Body
\toprule
Method & InfinityChat & Creative Writing & Image Gen. \\
\midrule
\multicolumn{4}{l}{\textbf{(A) Randomize over ideas, not tokens}} \\
\textbf{\ours (ours)} & $\bm{10.08}$ & $\bm{9.44}$ & $\bm{9.23}$ \\
Standard token sampling ($T{=}1$) & 2.41 & 5.25 & 4.01 \\
\quad Token sampling ($T{=}3$, $K{=}10$) & 3.28 & 3.07 & 4.66 \\
\quad Token sampling ($T{=}3$, $K{=}20$) & 3.46 & 0.44 & 4.93 \\
\quad Token sampling ($T{=}5$, $K{=}10$) & 3.19 & 0.31 & 4.90 \\
\quad Token sampling ($T{=}5$, $K{=}20$) & 2.15 & 0.03 & 4.99 \\
\midrule
\multicolumn{4}{l}{\textbf{(B) Sample by decoupling enumeration + selection}} \\
\textbf{\ours (ours)} & $\bm{10.61}$ & $\bm{10.08}$ & $\bm{8.21}$ \\
\quad No external RNG (LM selects choices from its list) & 5.34 & 6.75 & 5.46 \\
\quad No enumeration (directly prompt LM for decision) & 5.73 & 7.50 & 5.50 \\
\midrule
\multicolumn{4}{l}{\textbf{(C) Sample ideas as autoregressive local decisions}} \\
\textbf{\ours (ours)} & $\bm{10.61}$ & $\bm{10.08}$ & $\bm{8.21}$ \\
\quad Idea-list (prompt LM to list $k$ complete plans) & 7.61 & 8.20 & 7.54 \\
\quad Independent axes (no conditioning on past decisions) & 7.96 & 8.40 & 8.09 \\
\bottomrule
\end{NiceTabular}
\caption{
We compare \ours against ablations of our three design choices (\cref{sec:method}), measuring diversity as quality-gated Vendi score ($\uparrow$, $n$=128). We ablate \textbf{(A)} idea-space sampling via extreme-temperature top-$K$ token sampling, \textbf{(B)} external randomization over LM-enumerated candidates via no-RNG and no-enumeration variants, and \textbf{(C)} sampling ideas autoregressively as a series of conditional local decisions by instead generating complete ideas directly (idea-list) or sampling decisions independently (independent axes). All ablations drop diversity---often sharply. Results on (A) are from GPT-OSS-120B, where temperature is configurable; (B, C) show Opus 4.8.
}
\label{tab:ablations}
\end{table}

\section{Related Work}
\label{sec:related}
\textbf{(Lack of) language model diversity.} By now, a growing literature documents the limited diversity of LM generations on open-ended prompts~\citep{zhang2025noveltybench,kirk2023understanding,wu2025generative,xu2025echoes}, both within individual models and across model families~\citep{jiang2026artificial}.
This shortfall reduces user utility~\citep{zhang2025noveltybench}, homogenizes AI-assisted writing and decision-making~\citep{abdulhai2026llms,matz2025basic,padmakumar2024does}, and can constrain model development \citep{jung2026prismatic, cui2025entropy}, which often relies on diverse samples for e.g. reinforcement learning. Looking forward, limited diversity also limits open-ended scientific workflows~\citep{si2025can}. 
Improving diversity has been challenging because scaling and capability post-training tends to make models 
\textit{less} diverse~\citep{karouzos2026does,west2025base,murthy2025one}, as training often reduces token entropy. We show that this tension is not fundamental, leveraging the same IF capabilities that improve with post-training to drive greater diversity.

\textbf{Approaches for improving diversity.} Prior work has explored both training- and inference-time interventions. Training-time methods typically aim to make the LM token distribution itself more diverse, for instance by curating diverse training data or new post-training objectives~\citep{su2020diversifying,zhang2024forcing,springer2026annotations,shi2018toward,li2025jointly,sorensen2026spectrum,li2025preserving}. Inference-time methods have traditionally reshaped the sampling distribution~\citep{holtzman2019curious,roemmele2018automated}, and more recently steer generation toward different response modes by conditioning on seeds, personas, mode names, and other signals~\citep{misaki2026string,ge2024scaling,santurkar2023whose,wu2025mode}. Most related from this latter line of work is \citet{zhang2026you} which shows that diversifying plans diversifies generations, obtaining plans in a manner similar to our idea-list or independent-axes ablations of \ours. Our work crystallizes these recent efforts into an explicit instruction-following view of diversity, showing that external RNG combined with IF yields strong diversity without requiring LM token entropy.
\section{Conclusion}
In this work, we have shown that generation diversity can be treated as an instruction-following problem, shifting the locus of diversity away from LM token entropy---which can degrade with scale and post-training---toward a capability that improves as models do. Models need not encode all variation internally, but can instead act as reasoning components that convert external randomness from a surrounding system into useful, diverse responses. To do so, \ours relies on models being able to enumerate distinct options when instructed; we find that modern LMs already do well enough, allowing us to strongly outperform prior methods with just a simple harness. Natural next steps include optimizing models end-to-end with our harness, or distilling the resulting diverse generations back into the LM. Overall, we view IF-centric diversity not as a replacement for methods that improve the    model's intrinsic distribution, but as a complementary paradigm.
\section{Acknowledgements}
This project would not be possible without the time and support of a great number of others. We thank Victoria Graf for her thorough research and writing feedback throughout the project. We thank Melissa Hall, Reyhane Askari, and Yushi Hu for helpful discussions. We thank Ravi Arikapudi, Pedram Bashiri, Allen Bolourchi, and Mary Williamson for their executive support. 

\textbf{Funding.} This work was done while SG was part of the UW-Meta AI Mentorship program. SG is supported by the NSF GRFP. This work was also supported by the Singapore National Research Foundation and the National AI Group in the Singapore Ministry of Digital Development and Information under the AI Visiting Professorship Programme (award number AIVP-2024-001) and the AI2050 program at Schmidt Sciences.

\clearpage
\newpage
\bibliographystyle{assets/plainnat}
\bibliography{paper}

@article{kirk2023understanding,
  title={Understanding the effects of rlhf on llm generalisation and diversity},
  author={Kirk, Robert and Mediratta, Ishita and Nalmpantis, Christoforos and Luketina, Jelena and Hambro, Eric and Grefenstette, Edward and Raileanu, Roberta},
  journal={arXiv preprint arXiv:2310.06452},
  year={2023}
}

@article{zhang2025verbalized,
  title={Verbalized sampling: How to mitigate mode collapse and unlock llm diversity},
  author={Zhang, Jiayi and Yu, Simon and Chong, Derek and Sicilia, Anthony and Tomz, Michael R and Manning, Christopher D and Shi, Weiyan},
  journal={arXiv preprint arXiv:2510.01171},
  year={2025}
}

@article{paech2023eq,
  title={Eq-bench: An emotional intelligence benchmark for large language models},
  author={Paech, Samuel J},
  journal={arXiv preprint arXiv:2312.06281},
  year={2023}
}

@article{springer2026annotations,
  title={Annotations Mitigate Post-Training Mode Collapse},
  author={Springer, Jacob Mitchell and Advani, Madhu and Aichberger, Lukas and Bradley, Arwen and Malach, Eran and Saremi, Omid and Williamson, Sinead and Nakkiran, Preetum and Littwin, Etai and Raghunathan, Aditi},
  journal={arXiv preprint arXiv:2605.09995},
  year={2026}
}

@article{ahmed2025intent,
  title={Intent factored generation: Unleashing the diversity in your language model},
  author={Ahmed, Eltayeb and Berdica, Uljad and Elliott, Martha and Horak, Danijela and Foerster, Jakob N},
  journal={arXiv preprint arXiv:2506.09659},
  year={2025}
}

@article{jiang2026artificial,
  title={Artificial hivemind: The open-ended homogeneity of language models (and beyond)},
  author={Jiang, Liwei and Chai, Yuanjun and Li, Margaret and Liu, Mickel and Fok, Raymond and Dziri, Nouha and Tsvetkov, Yulia and Sap, Maarten and Choi, Yejin},
  journal={Advances in Neural Information Processing Systems},
  volume={38},
  year={2026}
}

@article{zhang2024forcing,
  title={Forcing diffuse distributions out of language models},
  author={Zhang, Yiming and Schwarzschild, Avi and Carlini, Nicholas and Kolter, Zico and Ippolito, Daphne},
  journal={arXiv preprint arXiv:2404.10859},
  year={2024}
}

@software{nvidia/Nemotron-Personas-USA,
  author = {Meyer, Yev and Corneil, Dane},
  title = {{Nemotron-Personas-USA}: Synthetic Personas Aligned to Real-World Distributions
},
  month = {June},
  year = {2025},
  url = {https://huggingface.co/datasets/nvidia/Nemotron-Personas-USA}
}

@article{holtzman2019curious,
  title={The curious case of neural text degeneration},
  author={Holtzman, Ari and Buys, Jan and Du, Li and Forbes, Maxwell and Choi, Yejin},
  journal={arXiv preprint arXiv:1904.09751},
  year={2019}
}

@article{li2025jointly,
  title={Jointly reinforcing diversity and quality in language model generations},
  author={Li, Tianjian and Zhang, Yiming and Yu, Ping and Saha, Swarnadeep and Khashabi, Daniel and Weston, Jason and Lanchantin, Jack and Wang, Tianlu},
  journal={arXiv preprint arXiv:2509.02534},
  year={2025}
}

@inproceedings{sorensen2026spectrum,
  title={Spectrum tuning: Post-training for distributional coverage and in-context steerability},
  author={Sorensen, Taylor and Newman, Benjamin and Moore, Jared and Park, Chan Young and Fisher, Jillian and Mireshghallah, Niloofar and Jiang, Liwei and Choi, Yejin},
  booktitle={International Conference on Learning Representations},
  volume={2026},
  pages={76138--76212},
  year={2026}
}

@article{zhang2026you,
  title={Where You Inject Diversity Matters: A Unified Framework for Diverse Generation},
  author={Zhang, Cheng and Xin, Rui and Zhong, Chudi},
  journal={arXiv preprint arXiv:2606.10302},
  year={2026}
}

@article{wu2025mode,
  title={Mode-conditioning unlocks superior test-time scaling},
  author={Wu, Chen Henry and Goyal, Sachin and Raghunathan, Aditi},
  journal={arXiv preprint arXiv:2512.01127},
  year={2025}
}

@inproceedings{misaki2026string,
  title={String seed of thought: Prompting LLMs for distribution-faithful and diverse generation},
  author={Misaki, Kou and Akiba, Takuya},
  booktitle={International Conference on Learning Representations},
  volume={2026},
  pages={89078--89136},
  year={2026}
}

@article{ge2024scaling,
  title={Scaling synthetic data creation with 1,000,000,000 personas},
  author={Ge, Tao and Chan, Xin and Wang, Xiaoyang and Yu, Dian and Mi, Haitao and Yu, Dong},
  journal={arXiv preprint arXiv:2406.20094},
  year={2024}
}

@article{zhang2025noveltybench,
  title={Noveltybench: Evaluating language models for humanlike diversity},
  author={Zhang, Yiming and Diddee, Harshita and Holm, Susan and Liu, Hanchen and Liu, Xinyue and Samuel, Vinay and Wang, Barry and Ippolito, Daphne},
  journal={arXiv preprint arXiv:2504.05228},
  year={2025}
}

@inproceedings{yao2019plan,
  title={Plan-and-write: Towards better automatic storytelling},
  author={Yao, Lili and Peng, Nanyun and Weischedel, Ralph and Knight, Kevin and Zhao, Dongyan and Yan, Rui},
  booktitle={Proceedings of the AAAI conference on artificial intelligence},
  volume={33},
  number={01},
  pages={7378--7385},
  year={2019}
}

@inproceedings{santurkar2023whose,
  title={Whose opinions do language models reflect?},
  author={Santurkar, Shibani and Durmus, Esin and Ladhak, Faisal and Lee, Cinoo and Liang, Percy and Hashimoto, Tatsunori},
  booktitle={International conference on machine learning},
  pages={29971--30004},
  year={2023},
  organization={PMLR}
}

@techreport{chatterji2025people,
  title={How people use chatgpt},
  author={Chatterji, Aaron and Cunningham, Thomas and Deming, David J and Hitzig, Zoe and Ong, Christopher and Shan, Carl Yan and Wadman, Kevin},
  year={2025},
  institution={National Bureau of Economic Research}
}

@inproceedings{
  zhao2024wildchat,
  title={WildChat: 1M Chat{GPT} Interaction Logs in the Wild},
  author={Wenting Zhao and Xiang Ren and Jack Hessel and Claire Cardie and Yejin Choi and Yuntian Deng},
  booktitle={The Twelfth International Conference on Learning Representations},
  year={2024},
  url={https://openreview.net/forum?id=Bl8u7ZRlbM}
}

@article{chang2026oneig,
  title={Oneig-bench: Omni-dimensional nuanced evaluation for image generation},
  author={Chang, Jingjing and Fang, Yixiao and Xing, Peng and Wu, Shuhan and Cheng, Wei and Wang, Rui and Zeng, Xianfang and Yu, Gang and Chen, Hai-Bao},
  journal={Advances in Neural Information Processing Systems},
  volume={38},
  year={2026}
}

@article{fu2023dreamsim,
  title={Dreamsim: Learning new dimensions of human visual similarity using synthetic data},
  author={Fu, Stephanie and Tamir, Netanel and Sundaram, Shobhita and Chai, Lucy and Zhang, Richard and Dekel, Tali and Isola, Phillip},
  journal={arXiv preprint arXiv:2306.09344},
  year={2023}
}

@article{friedman2022vendi,
  title={The vendi score: A diversity evaluation metric for machine learning},
  author={Friedman, Dan and Dieng, Adji Bousso},
  journal={arXiv preprint arXiv:2210.02410},
  year={2022}
}

@article{cui2025entropy,
  title={The entropy mechanism of reinforcement learning for reasoning language models},
  author={Cui, Ganqu and Zhang, Yuchen and Chen, Jiacheng and Yuan, Lifan and Wang, Zhi and Zuo, Yuxin and Li, Haozhan and Fan, Yuchen and Chen, Huayu and Chen, Weize and others},
  journal={arXiv preprint arXiv:2505.22617},
  year={2025}
}

@article{yakura2024empirical,
  title={Empirical evidence of Large Language Model's influence on human spoken communication},
  author={Yakura, Hiromu and Lopez-Lopez, Ezequiel and Brinkmann, Levin and de la Serna, Ignacio and Kirfel, Lara and Gupta, Prateek and Soraperra, Ivan and Eisenmann, Thomas F and Wulff, Dirk U and Rahwan, Iyad},
  journal={arXiv preprint arXiv:2409.01754},
  year={2024}
}

@article{abdulhai2026llms,
  title={How llms distort our written language},
  author={Abdulhai, Marwa and White, Isadora and Wan, Yanming and Qureshi, Ibrahim and Leibo, Joel and Kleiman-Weiner, Max and Jaques, Natasha},
  journal={arXiv preprint arXiv:2603.18161},
  year={2026}
}

@article{west2025base,
  title={Base models beat aligned models at randomness and creativity},
  author={West, Peter and Potts, Christopher},
  journal={arXiv preprint arXiv:2505.00047},
  year={2025}
}

@article{karouzos2026does,
  title={Where does output diversity collapse in post-training?},
  author={Karouzos, Constantinos and Tan, Xingwei and Aletras, Nikolaos},
  journal={arXiv preprint arXiv:2604.16027},
  year={2026}
}

@article{pyatkin2026generalizing,
  title={Generalizing verifiable instruction following},
  author={Pyatkin, Valentina and Malik, Saumya and Graf, Victoria and Ivison, Hamish and Huang, Shengyi and Dasigi, Pradeep and Lambert, Nathan and Hajishirzi, Hanna},
  journal={Advances in Neural Information Processing Systems},
  volume={38},
  year={2026}
}

@inproceedings{roemmele2018automated,
  title={Automated assistance for creative writing with an rnn language model},
  author={Roemmele, Melissa and Gordon, Andrew S},
  booktitle={Companion proceedings of the 23rd international conference on intelligent user interfaces},
  pages={1--2},
  year={2018}
}

@article{shi2018toward,
  title={Toward diverse text generation with inverse reinforcement learning},
  author={Shi, Zhan and Chen, Xinchi and Qiu, Xipeng and Huang, Xuanjing},
  journal={arXiv preprint arXiv:1804.11258},
  year={2018}
}

@inproceedings{su2020diversifying,
  title={Diversifying dialogue generation with non-conversational text},
  author={Su, Hui and Shen, Xiaoyu and Zhao, Sanqiang and Xiao, Zhou and Hu, Pengwei and Zhong, Randy and Niu, Cheng and Zhou, Jie},
  booktitle={Proceedings of the 58th Annual Meeting of the Association for Computational Linguistics},
  pages={7087--7097},
  year={2020}
}

@inproceedings{fan2018hierarchical,
  title={Hierarchical neural story generation},
  author={Fan, Angela and Lewis, Mike and Dauphin, Yann},
  booktitle={Proceedings of the 56th Annual Meeting of the Association for Computational Linguistics (Volume 1: Long Papers)},
  pages={889--898},
  year={2018}
}

@misc{wu2025qwenimagetechnicalreport,
      title={Qwen-Image Technical Report},
      author={Chenfei Wu and Jiahao Li and Jingren Zhou and Junyang Lin and Kaiyuan Gao and Kun Yan and Sheng-ming Yin and Shuai Bai and Xiao Xu and Yilei Chen and Yuxiang Chen and Zecheng Tang and Zekai Zhang and Zhengyi Wang and An Yang and Bowen Yu and Chen Cheng and Dayiheng Liu and Deqing Li and Hang Zhang and Hao Meng and Hu Wei and Jingyuan Ni and Kai Chen and Kuan Cao and Liang Peng and Lin Qu and Minggang Wu and Peng Wang and Shuting Yu and Tingkun Wen and Wensen Feng and Xiaoxiao Xu and Yi Wang and Yichang Zhang and Yongqiang Zhu and Yujia Wu and Yuxuan Cai and Zenan Liu},
      year={2025},
      eprint={2508.02324},
      archivePrefix={arXiv},
      primaryClass={cs.CV},
      url={https://arxiv.org/abs/2508.02324},
}

@article{watson2023novo,
  title={De novo design of protein structure and function with RFdiffusion},
  author={Watson, Joseph L and Juergens, David and Bennett, Nathaniel R and Trippe, Brian L and Yim, Jason and Eisenach, Helen E and Ahern, Woody and Borst, Andrew J and Ragotte, Robert J and Milles, Lukas F and others},
  journal={Nature},
  volume={620},
  number={7976},
  pages={1089--1100},
  year={2023},
  publisher={Nature Publishing Group UK London}
}

@article{listov2024opportunities,
  title={Opportunities and challenges in design and optimization of protein function},
  author={Listov, Dina and Goverde, Casper A and Correia, Bruno E and Fleishman, Sarel Jacob},
  journal={Nature Reviews Molecular Cell Biology},
  volume={25},
  number={8},
  pages={639--653},
  year={2024},
  publisher={Nature Publishing Group UK London}
}

@article{van2024fast,
  title={Fast and accurate protein structure search with Foldseek},
  author={Van Kempen, Michel and Kim, Stephanie S and Tumescheit, Charlotte and Mirdita, Milot and Lee, Jeongjae and Gilchrist, Cameron LM and S{\"o}ding, Johannes and Steinegger, Martin},
  journal={Nature biotechnology},
  volume={42},
  number={2},
  pages={243--246},
  year={2024},
  publisher={Nature Publishing Group US New York}
}

@article{zhang2004scoring,
  title={Scoring function for automated assessment of protein structure template quality},
  author={Zhang, Yang and Skolnick, Jeffrey},
  journal={Proteins: Structure, Function, and Bioinformatics},
  volume={57},
  number={4},
  pages={702--710},
  year={2004},
  publisher={Wiley Online Library}
}

@inproceedings{wu2025generative,
  title={Generative monoculture in large language models},
  author={Wu, Fan and Black, Emily and Chandrasekaran, Varun},
  booktitle={International Conference on Learning Representations},
  volume={2025},
  pages={33068--33107},
  year={2025}
}

@inproceedings{padmakumar2024does,
  title={Does writing with language models reduce content diversity?},
  author={Padmakumar, Vishakh and He, He},
  booktitle={International Conference on Learning Representations},
  volume={2024},
  pages={642--669},
  year={2024}
}

@article{xu2025echoes,
  title={Echoes in ai: Quantifying lack of plot diversity in llm outputs},
  author={Xu, Weijia and Jojic, Nebojsa and Rao, Sudha and Brockett, Chris and Dolan, Bill},
  journal={Proceedings of the National Academy of Sciences},
  volume={122},
  number={35},
  pages={e2504966122},
  year={2025},
  publisher={National Academy of Sciences}
}

@article{matz2025basic,
  title={The Basic B*** Effect: The Use of LLM-based Agents Reduces the Distinctiveness and Diversity of People's Choices},
  author={Matz, Sandra C and Horton, C Blaine and Goethals, Sofie},
  journal={arXiv preprint arXiv:2509.02910},
  year={2025}
}

@misc{qwen3.5,
    title  = {{Qwen3.5}: Towards Native Multimodal Agents},
    author = {{Qwen Team}},
    month  = {February},
    year   = {2026},
    url    = {https://qwen.ai/blog?id=qwen3.5}
}

@article{agarwal2025gpt,
  title={gpt-oss-120b \& gpt-oss-20b model card},
  author={Agarwal, Sandhini and Ahmad, Lama and Ai, Jason and Altman, Sam and Applebaum, Andy and Arbus, Edwin and Arora, Rahul K and Bai, Yu and Baker, Bowen and Bao, Haiming and others},
  journal={arXiv preprint arXiv:2508.10925},
  year={2025}
}

@article{team2026gemma,
  title={Gemma 4 technical report},
  author={Team, Gemma and Abd, Sherif El and Aggarwal, Vaibhav and Algayres, Robin and Andreev, Alek and Bachem, Olivier and Ballantyne, Ian and Brick, Cormac and C{\u{a}}rbune, Victor and Casbon, Michelle and others},
  journal={arXiv preprint arXiv:2607.02770},
  year={2026}
}

@misc{opus48,
  author       = {{Anthropic}},
  title        = {Introducing Claude Opus 4.8},
  year         = {2026},
  month        = may,
  howpublished = {\url{https://www.anthropic.com/news/claude-opus-4-8}},
  note         = {System card: \url{https://www.anthropic.com/claude-opus-4-8-system-card}. Model version \texttt{claude-opus-4-8}}
}

@misc{gpt55,
  author       = {{OpenAI}},
  title        = {Introducing GPT-5.5},
  year         = {2026},
  month        = apr,
  howpublished = {\url{https://openai.com/index/introducing-gpt-5-5/}},
  note         = {Model version \texttt{gpt-5.5}}
}

@misc{inkling,
  author       = {{Thinking Machines Lab}},
  title        = {Inkling: Our Open-Weights Model},
  year         = {2026},
  month        = jul,
  howpublished = {\url{https://thinkingmachines.ai/news/introducing-inkling/}},
  note         = {Inkling-Small: \url{https://thinkingmachines.ai/news/inkling-small/}}
}

@article{shypula2025evaluating,
  title={Evaluating the diversity and quality of llm generated content},
  author={Shypula, Alexander and Li, Shuo and Zhang, Botong and Padmakumar, Vishakh and Yin, Kayo and Bastani, Osbert},
  journal={arXiv preprint arXiv:2504.12522},
  year={2025}
}

@article{cao2022design,
  title={Design of protein-binding proteins from the target structure alone},
  author={Cao, Longxing and Coventry, Brian and Goreshnik, Inna and Huang, Buwei and Sheffler, William and Park, Joon Sung and Jude, Kevin M and Markovi{\'c}, Iva and Kadam, Rameshwar U and Verschueren, Koen HG and others},
  journal={Nature},
  volume={605},
  number={7910},
  pages={551--560},
  year={2022},
  publisher={Nature Publishing Group UK London}
}

@article{pacesa2025one,
  title={One-shot design of functional protein binders with BindCraft},
  author={Pacesa, Martin and Nickel, Lennart and Schellhaas, Christian and Schmidt, Joseph and Pyatova, Ekaterina and Kissling, Lucas and Barendse, Patrick and Choudhury, Jagrity and Kapoor, Srajan and Alcaraz-Serna, Ana and others},
  journal={Nature},
  volume={646},
  number={8084},
  pages={483--492},
  year={2025},
  publisher={Nature Publishing Group UK London}
}

@article{candido2026language,
  title={Language modeling materializes a world model of protein biology},
  author={Candido, Salvatore and Hayes, Thomas and Derry, Alexander and Rao, Roshan and Lin, Zeming and Verkuil, Robert and Wu, Bryan Z and Lee, Jin Sub and Bruguera, Elise S and Keval, Jehan A and others},
  journal={bioRxiv},
  pages={2026--06},
  year={2026},
  publisher={Cold Spring Harbor Laboratory}
}

@article{dunbrack2025res,
  title={R{\=e}s ipSAE loquunt: What’s wrong with AlphaFold’s ipTM score and how to fix it},
  author={Dunbrack Jr, Roland L},
  journal={Biorxiv},
  year={2025}
}

@misc{shanehsazzadeh-autonomous-protein-binder-design-2026, author = {Shanehsazzadeh, Amir}, title = {Autonomous De Novo Protein Binder Design with Claude}, publisher = {arXiv}, year = {2026}}

@misc{lee2026prompteina,
  author = {Lee, Joseph and Li, Yanjing and Qiang, Bo and Lau, Gregory Kang Ruey and Ettinger, Allyson and Smith, Noah A. and Donovan-Maiye, Rory and Baker, David and Hajishirzi, Hannaneh and Koh, Pang Wei},
  title  = {Designing Protein Structures from Natural Language},
  year   = {2026},
  note   = {Forward cite for a forthcoming model and paper release.},
}

@article{zhang2025qwen3,
  title={Qwen3 embedding: Advancing text embedding and reranking through foundation models},
  author={Zhang, Yanzhao and Li, Mingxin and Long, Dingkun and Zhang, Xin and Lin, Huan and Yang, Baosong and Xie, Pengjun and Yang, An and Liu, Dayiheng and Lin, Junyang and others},
  journal={arXiv preprint arXiv:2506.05176},
  year={2025}
}

@article{zheng2024sglang,
  title={Sglang: Efficient execution of structured language model programs},
  author={Zheng, Lianmin and Yin, Liangsheng and Xie, Zhiqiang and Sun, Chuyue and Huang, Jeff and Yu, Cody H and Cao, Shiyi and Kozyrakis, Christos and Stoica, Ion and Gonzalez, Joseph E and others},
  journal={Advances in neural information processing systems},
  volume={37},
  pages={62557--62583},
  year={2024}
}

@inproceedings{si2025can,
  title={Can llms generate novel research ideas? a large-scale human study with 100+ nlp researchers},
  author={Si, Chenglei and Yang, Diyi and Hashimoto, Tatsunori},
  booktitle={International Conference on Learning Representations},
  volume={2025},
  pages={94003--94092},
  year={2025}
}

@article{jung2026prismatic,
  title={Prismatic synthesis: Gradient-based data diversification boosts generalization in llm reasoning},
  author={Jung, Jaehun and Han, Seungju and Lu, Ximing and Hallinan, Skyler and Acuna, David and Prabhumoye, Shrimai and Patwary, Mostofa and Shoeybi, Mohammad and Catanzaro, Bryan and Choi, Yejin},
  journal={Advances in Neural Information Processing Systems},
  volume={38},
  pages={90649--90685},
  year={2026}
}

@inproceedings{murthy2025one,
  title={One fish, two fish, but not the whole sea: Alignment reduces language models’ conceptual diversity},
  author={Murthy, Sonia Krishna and Ullman, Tomer and Hu, Jennifer},
  booktitle={Proceedings of the 2025 Conference of the Nations of the Americas Chapter of the Association for Computational Linguistics: Human Language Technologies (Volume 1: Long Papers)},
  pages={11241--11258},
  year={2025}
}

@inproceedings{li2025preserving,
  title={Preserving diversity in supervised fine-tuning of large language models},
  author={Li, Ziniu and Chen, Congliang and Xu, Tian and Qin, Zeyu and Xiao, Jiancong and Luo, Zhi-Quan and Sun, Ruoyu},
  booktitle={International Conference on Learning Representations},
  volume={2025},
  pages={66127--66154},
  year={2025}
}

\clearpage
\newpage
\beginappendix

\crefalias{section}{appendix}
\crefalias{subsection}{subappendix}
\crefalias{subsubsection}{subsubappendix}

\section{Extended results}
\label{appdx:results}

\subsection{Qualitative examples}
\label{appdx:results:examples}
See \cref{fig:story_qualitative} for qualitative story examples generated by \ours and baselines.

\newcommand{\pov}[1]{{\bfseries #1}}   

\begin{figure}[htbp]
\centering
\begin{storybox}{storybase}{Direct sampling --- third person}
The flyer in the elevator promised everything \pov{Mei} wanted: a two-bedroom
HDB flat in Tampines, resale price forty thousand below market. \pov{She}
should have known better. The auntie who sold it barely made eye contact.
``Renovate quick,'' she said, pressing the keys into \pov{Mei's} palm. ``Don't
sleep in the small room first.'' \pov{Mei} laughed it off. Superstition. Every
cheap flat in Singapore came with a story. You didn't get discounts for
nothing. The first week was fine. Then came the knocking. Not on the door. From
inside the walls of the small bedroom---three slow raps, always at 3:07 AM.
\pov{She} checked the pipes, the aircon, the neighbors. Nothing. By the second
week, \pov{she} smelled it: joss sticks, thick and sweet, though \pov{she}
burned none. [\ldots]
\end{storybox}

\begin{storybox}{storybase}{Verbalized Sampling --- third person}
\pov{Ravi} bought the resale flat because it was cheap---suspiciously cheap for
a five-room unit in Sengkang. The agent had smiled too much, talked too fast,
and never once mentioned why the previous family had left after only two
months. The first strange thing was the humming. It came from the master
bedroom wardrobe, a low tuneless drone that \pov{Ravi} first mistook for a
fault in the aircon. But when \pov{he} pressed \pov{his} ear to the wooden
doors, the humming stopped, replaced by breathing---slow, patient, waiting.
\pov{He} found the offerings on the third night: small piles of rice arranged
in careful spirals, though \pov{he} had emptied the space \pov{himself}.
[\ldots]
\end{storybox}

\begin{storybox}{metablue}{\ours --- second person}
\pov{You} move into Block 114 because it is cheap, and it is cheap because it
is dying. The agent doesn't say this. She says \emph{quiet}, she says
\emph{character}, she says \emph{the en-bloc talks fell through so prices are
soft}. But \pov{you} count the dark windows as \pov{you} crane \pov{your} neck
upward in the void deck, and \pov{you} understand. Of the fifty-odd units
stacked above \pov{you}, perhaps six are lit. The rest are hollow squares,
their families long gone to newer estates, leaving behind only the smell of old
water and the ghost of a thousand dinners cooked in the same narrow kitchens.
\pov{You} tell \pov{yourself} the loneliness is temporary. The water tastes
wrong from the first night. Not foul---\emph{full}. It sits on \pov{your}
tongue like something with a memory. \pov{Your} hand turns the faucet before
\pov{your} mind consents. [\ldots]
\end{storybox}

\caption{\ours can discover novel modes, surfacing several stories narrated in
the second person when no other approach does so across 256 samples. Pronouns
are bolded for emphasis. The task prompt, ``Write me a short horror story about
singapore apartment house,'' is drawn from WildChat~\citep{zhao2024wildchat}
and included amongst our story evaluation prompts.}
\label{fig:story_qualitative}
\end{figure}

\subsection{Additional main results}
\label{appdx:results:main}
See \cref{tab:main_results_full} for the concrete quality and diversity metrics numbers from our main experiments.

\colorlet{gatedtint}{metablue!10}
\providecommand{\rawv}[1]{{\fontsize{5.5pt}{6.5pt}\selectfont\color{gray!80!black}(#1)}}
\begin{table*}[htbp]
\centering
{\fontsize{6.7pt}{7.7pt}\selectfont
\setlength{\tabcolsep}{2.0pt}
\renewcommand{\arraystretch}{1.10}
\begin{NiceTabular}{@{}l rrr rrr rrr@{}}
\CodeBefore
  \rectanglecolor{gatedtint}{2-2}{33-2}
  \rectanglecolor{gatedtint}{2-5}{33-5}
  \rectanglecolor{gatedtint}{2-8}{33-8}
  \rowcolor{metabg}{3,10,17,26}
\Body
\toprule
& \multicolumn{3}{c}{\textbf{InfinityChat}}
& \multicolumn{3}{c}{\textbf{Creative Writing}}
& \multicolumn{3}{c}{\textbf{Image Generation}} \\
\cmidrule(lr){2-4} \cmidrule(lr){5-7} \cmidrule(lr){8-10}
\textbf{Method}
& \shortstack[c]{Quality-\\gated div.\\(raw div.)} & \shortstack[c]{Avg.\ quality\\of passing\\(raw qual.)} & \shortstack[c]{Gate\\pass\\rate (\%)}
& \shortstack[c]{Quality-\\gated div.\\(raw div.)} & \shortstack[c]{Avg.\ quality\\of passing\\(raw qual.)} & \shortstack[c]{Gate\\pass\\rate (\%)}
& \shortstack[c]{Quality-\\gated div.\\(raw div.)} & \shortstack[c]{Avg.\ quality\\of passing\\(raw qual.)} & \shortstack[c]{Gate\\pass\\rate (\%)} \\
\midrule
\multicolumn{10}{l}{\textbf{Claude Opus 4.8}} \\
\textbf{Gacha (ours)}
& $\bm{13.12}$ \rawv{15.14} & 1.00 \rawv{0.80} & 79.5
& $\bm{11.84}$ \rawv{11.92} & 17.4 \rawv{17.3} & 97.8
& $\bm{10.03}$ \rawv{13.37} & 0.94 \rawv{0.82} & 57.6 \\
List
& 7.87 \rawv{8.62} & 1.00 \rawv{0.89} & 88.7
& 5.65 \rawv{9.67} & 14.9 \rawv{11.8} & 19.0
& 6.78 \rawv{7.89} & 0.95 \rawv{0.85} & 66.2 \\
Verbalized
& 4.10 \rawv{4.19} & 1.00 \rawv{0.95} & 95.4
& 6.91 \rawv{8.31} & 15.3 \rawv{14.1} & 46.8
& 3.58 \rawv{3.92} & 0.96 \rawv{0.87} & 67.6 \\
Persona
& 5.54 \rawv{5.67} & 1.00 \rawv{0.94} & 93.8
& 8.37 \rawv{8.84} & 16.4 \rawv{15.7} & 83.8
& 4.63 \rawv{5.04} & 0.95 \rawv{0.87} & 74.9 \\
Seed
& 2.89 \rawv{2.90} & 1.00 \rawv{0.98} & 97.7
& 6.14 \rawv{6.16} & 16.6 \rawv{16.5} & 97.0
& 4.07 \rawv{4.43} & 0.95 \rawv{0.86} & 73.1 \\
Direct ($T{=}1$)
& 1.84 \rawv{1.85} & 1.00 \rawv{0.95} & 94.8
& 5.53 \rawv{5.62} & 15.9 \rawv{15.2} & 79.5
& 2.98 \rawv{3.11} & 0.94 \rawv{0.88} & 76.4 \\
\midrule
\multicolumn{10}{l}{\textbf{GPT-5.5}} \\
\textbf{Gacha (ours)}
& $\bm{9.87}$ \rawv{10.61} & 1.00 \rawv{0.90} & 89.6
& $\bm{11.10}$ \rawv{11.13} & 17.5 \rawv{17.5} & 98.9
& $\bm{7.29}$ \rawv{8.36} & 0.95 \rawv{0.85} & 68.7 \\
List
& 7.19 \rawv{7.50} & 1.00 \rawv{0.91} & 91.3
& 6.63 \rawv{8.70} & 15.2 \rawv{13.7} & 32.4
& 5.86 \rawv{6.56} & 0.94 \rawv{0.86} & 69.9 \\
Verbalized
& 5.11 \rawv{5.15} & 1.00 \rawv{0.96} & 96.0
& 8.18 \rawv{8.71} & 16.0 \rawv{15.6} & 73.9
& 3.78 \rawv{4.04} & 0.95 \rawv{0.87} & 72.9 \\
Persona
& 3.22 \rawv{3.22} & 1.00 \rawv{0.98} & 98.2
& 6.73 \rawv{7.09} & 16.0 \rawv{15.6} & 80.6
& 3.21 \rawv{3.35} & 0.95 \rawv{0.88} & 77.2 \\
Seed
& 2.05 \rawv{2.06} & 1.00 \rawv{0.98} & 98.2
& 5.14 \rawv{5.21} & 16.0 \rawv{15.7} & 87.9
& 2.94 \rawv{3.04} & 0.95 \rawv{0.88} & 77.1 \\
Direct ($T{=}1$)
& 1.94 \rawv{1.94} & 1.00 \rawv{0.98} & 97.5
& 5.27 \rawv{5.37} & 16.0 \rawv{15.5} & 82.6
& 2.84 \rawv{2.94} & 0.95 \rawv{0.88} & 77.6 \\
\midrule
\multicolumn{10}{l}{\textbf{GPT-OSS 120B}} \\
\textbf{Gacha (ours)}
& $\bm{15.01}$ \rawv{19.16} & 1.00 \rawv{0.64} & 63.6
& $\bm{12.84}$ \rawv{13.73} & 14.1 \rawv{13.4} & 77.1
& $\bm{13.85}$ \rawv{17.09} & 0.92 \rawv{0.82} & 61.0 \\
List
& 10.27 \rawv{12.57} & 1.00 \rawv{0.76} & 76.0
& 3.90 \rawv{10.86} & 11.4 \rawv{6.8} & 6.1
& 10.76 \rawv{12.40} & 0.93 \rawv{0.84} & 66.9 \\
Verbalized
& 7.49 \rawv{8.61} & 1.00 \rawv{0.80} & 79.8
& 6.43 \rawv{9.11} & 12.8 \rawv{10.6} & 25.4
& 6.10 \rawv{6.80} & 0.93 \rawv{0.85} & 71.1 \\
Persona
& 5.53 \rawv{5.70} & 1.00 \rawv{0.89} & 89.1
& 8.95 \rawv{10.02} & 13.7 \rawv{12.1} & 57.2
& 5.81 \rawv{6.11} & 0.93 \rawv{0.87} & 77.6 \\
Seed
& 3.77 \rawv{3.98} & 1.00 \rawv{0.87} & 87.3
& 6.42 \rawv{7.11} & 13.4 \rawv{11.6} & 47.3
& 6.18 \rawv{6.55} & 0.93 \rawv{0.86} & 75.5 \\
Direct Planning
& 3.42 \rawv{3.47} & 1.00 \rawv{0.93} & 93.4
& 5.94 \rawv{6.27} & 13.6 \rawv{12.5} & 65.0
& 4.93 \rawv{5.11} & 0.93 \rawv{0.87} & 78.6 \\
Direct ($T{=}1.5$)
& 3.06 \rawv{3.11} & 1.00 \rawv{0.89} & 89.5
& 6.32 \rawv{7.06} & 13.6 \rawv{11.5} & 52.1
& 5.19 \rawv{5.43} & 0.93 \rawv{0.87} & 79.2 \\
Direct ($T{=}1$)
& 2.62 \rawv{2.64} & 1.00 \rawv{0.92} & 92.4
& 6.26 \rawv{6.45} & 13.8 \rawv{12.8} & 74.5
& 4.85 \rawv{5.04} & 0.93 \rawv{0.87} & 79.4 \\
\midrule
\multicolumn{10}{l}{\textbf{Inkling Large}} \\
\textbf{Gacha (ours)}
& $\bm{17.31}$ \rawv{20.78} & 1.00 \rawv{0.73} & 72.7
& $\bm{12.12}$ \rawv{13.59} & 17.0 \rawv{16.0} & 52.2
& $\bm{13.40}$ \rawv{16.61} & 0.93 \rawv{0.83} & 58.1 \\
List
& 13.69 \rawv{16.21} & 1.00 \rawv{0.77} & 77.2
& 3.71 \rawv{10.96} & 16.0 \rawv{11.7} & 3.5
& 11.50 \rawv{13.86} & 0.93 \rawv{0.84} & 61.0 \\
Verbalized
& 4.15 \rawv{4.45} & 1.00 \rawv{0.89} & 89.1
& 6.71 \rawv{8.24} & 16.8 \rawv{15.8} & 43.9
& 4.20 \rawv{4.54} & 0.94 \rawv{0.87} & 74.1 \\
Persona
& 8.55 \rawv{9.08} & 1.00 \rawv{0.85} & 85.2
& 8.95 \rawv{10.54} & 16.8 \rawv{15.8} & 48.2
& 6.33 \rawv{6.79} & 0.94 \rawv{0.87} & 74.3 \\
Seed
& 4.69 \rawv{4.96} & 1.00 \rawv{0.86} & 86.3
& 6.35 \rawv{7.09} & 16.9 \rawv{15.6} & 44.1
& 5.92 \rawv{6.29} & 0.94 \rawv{0.87} & 75.3 \\
Direct Planning
& 3.60 \rawv{3.64} & 1.00 \rawv{0.95} & 95.3
& 6.00 \rawv{6.91} & 16.8 \rawv{14.5} & 39.6
& 4.24 \rawv{4.40} & 0.94 \rawv{0.88} & 77.6 \\
Direct ($T{=}1$)
& 2.75 \rawv{2.76} & 1.00 \rawv{0.98} & 97.5
& 6.39 \rawv{6.47} & 17.3 \rawv{17.0} & 88.8
& 3.79 \rawv{3.91} & 0.94 \rawv{0.88} & 80.0 \\
\bottomrule
\end{NiceTabular}
}
\caption{
Extended main results for our experiments in \cref{sec:mainexps}, showing the concrete quality and diversity scores underlying \cref{fig:mainresult}. \textbf{Quality-gated diversity}: Vendi score over the samples that pass a quality gate, our main metric. We show raw Vendi over all samples---including low-quality ones---in parentheses. \textbf{Average quality of passing samples}: mean judge quality of the passing samples (chat response validity, binary 0/1; story rubric-judged quality, 0--20; image rubric-judged alignment, 0--1), with the mean over all samples in parentheses. \textbf{Quality-gate pass rate}: the percentage of samples that pass the quality gate, defined to be $0.95\times$ the model's direct-sampling mean on the same prompt. We show results at the highest number of samples per prompt considered for each model ($n{=}512$ for open-weight models GPT-OSS and Inkling, and $n{=}256$ on closed models). 
}
\label{tab:main_results_full}
\end{table*}

\subsection{Diversity under greedy decoding} 
\label{appdx:results:greedy}

\definecolor{metared}{HTML}{E4405F}
\newcommand{\neutral}{\cellcolor{metafg!15}}
\newcommand{\dropI}{\cellcolor{metared!12}}
\newcommand{\dropII}{\cellcolor{metared!25}}
\newcommand{\dropIV}{\cellcolor{metared!50}}

\begin{table*}[!ht]
\centering
{\fontsize{9pt}{11pt}\selectfont
\setlength{\tabcolsep}{3.75pt}
\renewcommand{\arraystretch}{1.10}
\begin{NiceTabular}{@{}l rrr rrr rrr@{}}
\CodeBefore
  \rowcolor{metabg}{3,9,15}
\Body
\toprule
& \multicolumn{3}{c}{\textbf{InfinityChat}}
& \multicolumn{3}{c}{\textbf{Creative Writing}}
& \multicolumn{3}{c}{\textbf{Image Generation}} \\
\cmidrule(lr){2-4} \cmidrule(lr){5-7} \cmidrule(lr){8-10}
\textbf{Method}
& $T{=}1$ & Greedy & $\Delta$
& $T{=}1$ & Greedy & $\Delta$
& $T{=}1$ & Greedy & $\Delta$ \\
\midrule
\multicolumn{10}{l}{\textbf{GPT-OSS 120B}} \\
\textbf{\ours (ours)}
& $\bm{10.08}$ & $\bm{9.62}$ & \neutral{$-$4.6\%}
& $\bm{9.44}$ & $\bm{9.20}$ & \neutral{$-$2.5\%}
& $\bm{9.23}$ & $\bm{8.82}$ & \neutral{$-$4.5\%} \\
List
& 7.62 & 4.71 & \dropII{$-$38.1\%}
& 1.93 & 0.06 & \dropIV{$-$96.8\%}
& 7.84 & 4.48 & \dropII{$-$42.9\%} \\
Verbalized
& 6.07 & 3.83 & \dropII{$-$36.9\%}
& 4.49 & 0.65 & \dropIV{$-$85.5\%}
& 4.73 & 3.19 & \dropII{$-$32.7\%} \\
Persona
& 4.64 & 4.64 & \neutral{$+$0.2\%}
& 6.71 & 6.58 & \neutral{$-$1.8\%}
& 4.74 & 4.67 & \neutral{$-$1.5\%} \\
Seed
& 3.35 & 3.08 & \dropI{$-$8.0\%}
& 5.12 & 4.74 & \dropI{$-$7.3\%}
& 4.95 & 4.68 & \dropI{$-$5.5\%} \\
\midrule
\multicolumn{10}{l}{\textbf{Inkling Large}} \\
\textbf{\ours (ours)}
& $\bm{11.65}$ & $\bm{11.65}$ & \neutral{0.0\%}
& $\bm{8.44}$ & $\bm{8.03}$ & \neutral{$-$4.8\%}
& $\bm{9.04}$ & $\bm{9.05}$ & \neutral{$+$0.1\%} \\
List
& 10.11 & 5.90 & \dropII{$-$41.7\%}
& 1.54 & 0.06 & \dropIV{$-$96.0\%}
& 8.07 & 4.94 & \dropII{$-$38.8\%} \\
Verbalized
& 3.96 & 1.69 & \dropIV{$-$57.4\%}
& 6.04 & 0.54 & \dropIV{$-$91.0\%}
& 4.06 & 1.23 & \dropIV{$-$69.6\%} \\
Persona
& 6.65 & 6.56 & \neutral{$-$1.4\%}
& 6.54 & 6.07 & \dropI{$-$7.1\%}
& 5.07 & 4.90 & \neutral{$-$3.2\%} \\
Seed
& 4.00 & 3.85 & \neutral{$-$3.9\%}
& 4.93 & 4.45 & \dropI{$-$9.9\%}
& 4.76 & 4.64 & \neutral{$-$2.6\%} \\
\midrule
\multicolumn{10}{l}{\textbf{Qwen3.5-27B}} \\
\textbf{\ours (ours)}
& $\bm{11.29}$ & $\bm{10.62}$ & \dropI{$-$6.0\%}
& $\bm{9.29}$ & $\bm{8.37}$ & \dropI{$-$9.9\%}
& $\bm{8.61}$ & $\bm{8.31}$ & \neutral{$-$3.5\%} \\
List
& 11.01 & 5.29 & \dropIV{$-$51.9\%}
& 2.67 & 0.00 & \dropIV{$-$100.0\%}
& 8.42 & 4.64 & \dropII{$-$44.8\%} \\
Verbalized
& 7.64 & 4.31 & \dropII{$-$43.6\%}
& 1.84 & 0.00 & \dropIV{$-$100.0\%}
& 5.99 & 3.57 & \dropII{$-$40.4\%} \\
Persona
& 7.92 & 7.38 & \dropI{$-$6.8\%}
& 6.87 & 6.18 & \dropII{$-$10.0\%}
& 5.40 & 5.06 & \dropI{$-$6.3\%} \\
Seed
& 4.60 & 3.54 & \dropII{$-$23.0\%}
& 4.64 & 4.00 & \dropII{$-$13.9\%}
& 4.91 & 4.07 & \dropII{$-$17.1\%} \\
\bottomrule
\end{NiceTabular}
}
\caption{We show quality-gated Vendi score diversity ($\uparrow$ higher is better, $n{=}128$) with \ours and other inference-time methods under (a) standard token sampling ($T{=}1$) and (b) greedy decoding. Because \ours externalizes randomness from the LM and does not rely on token entropy, switching to greedy decoding only negligibly affects the diversity it elicits ($\Delta$). The near-complete drop in Verbalized Sampling and list prompting diversity on creative writing occurs because these methods often degenerate into nonsensical responses under greedy decoding, such that few if any responses pass the quality gate.}
\label{tab:greedy_decoding}
\end{table*}

See \cref{tab:greedy_decoding} for results comparing \ours and baselines under standard token sampling (temperature=1.0) and greedy decoding. Overall, \ours retains its diversity while list-based methods drop severely. Persona and seed prompting also retain much of their diversity because they also externalize some variation from the LM. \ours is the only method that combines strong diversity with independence from token entropy.

\subsection{Additional ablation results}
\label{appdx:results:ablations}

\begin{table}[htbp]
\centering
\setlength{\tabcolsep}{6pt}
\begin{NiceTabular}{lccc}
\CodeBefore
  \rowcolor{metabg}{2,9,13}
\Body
\toprule
Method & InfinityChat & Creative Writing & Image Gen. \\
\midrule
\multicolumn{4}{l}{\textbf{(A) Randomize over ideas, not tokens}} \\
\textbf{\ours (ours)} & $\bm{11.65}$ & $\bm{8.44}$ & $\bm{9.04}$ \\
Standard token sampling ($T{=}1$) & 2.54 & 5.45 & 3.26 \\
\quad Token sampling ($T{=}3$, $K{=}10$) & 3.07 & 2.56 & 4.25 \\
\quad Token sampling ($T{=}3$, $K{=}20$) & 2.16 & 0.00 & 4.90 \\
\quad Token sampling ($T{=}5$, $K{=}10$) & 1.79 & 0.03 & 5.02 \\
\quad Token sampling ($T{=}5$, $K{=}20$) & 0.42 & 0.00 & 5.29 \\
\midrule
\multicolumn{4}{l}{\textbf{(B) Sample by decoupling enumeration + selection}} \\
\textbf{\ours (ours)} & $\bm{8.24}$ & $\bm{9.53}$ & $\bm{6.25}$ \\
\quad No external RNG (LM selects choices from its list) & 3.95 & 6.41 & 5.08 \\
\quad No enumeration (directly prompt LM for decision) & 4.58 & 6.38 & 4.42 \\
\midrule
\multicolumn{4}{l}{\textbf{(C) Sample ideas as autoregressive local decisions}} \\
\textbf{\ours (ours)} & $\bm{8.24}$ & $\bm{9.53}$ & $\bm{6.25}$ \\
\quad Idea-list (prompt LM to list $k$ complete plans) & 5.96 & 7.95 & 5.43 \\
\quad Independent axes (no conditioning on past decisions) & 6.34 & 8.20 & 6.02 \\
\bottomrule
\end{NiceTabular}
\caption{
Additional ablation results to corroborate~\cref{tab:ablations}. Panel (A) shows results with Inkling Large; panels (B) and (C) show GPT-5.5.
}
\label{tab:more_ablations}
\end{table}

We provide additional discussion and corroborating results for the main ablations shown in \cref{tab:ablations}. Each ablation removes one of the three design choices underlying \ours: (A) randomizing over semantic ideas rather than tokens, (B) decoupling enumeration from selection with external RNG, and (C) sampling ideas autoregressively. Our main ablations table showed results on GPT-OSS for (A), as it allows configuring sampling parameters, and results on Opus 4.8 for (B, C). Here, we show corroborating results on Inkling and GPT 5.5 (\cref{tab:more_ablations}).

\paragraph{(A) Idea-space vs.\ token-space randomization.}
\ours works in idea space, aggressively randomizing over curated sets of LM-generated ideas by using external RNG to uniformly sample. To test whether the same strategy can be applied at the token level, we compare against extreme-temperature token sampling with top-$p$~\citep{holtzman2019curious} and top-$K$~\citep{fan2018hierarchical} decoding. At each decoding step, top-$K$ leverages the LM to obtain an analogous ``curated set'' of most likely next tokens; increasing temperature makes sampling over this set progressively more uniform. We evaluate several values of $K$ and temperature, fixing $p=0.95$. We find that even for relatively small $K$, increasing temperature beyond 3---perhaps unsurprisingly---greatly damages the coherence of responses and consequently limits quality-filtered diversity. Note that temperature 3 is still far from uniform sampling, which corresponds to the infinite temperature limit. Working in idea space is critical for the aggressive randomness that enables \ours.

\paragraph{(B) Decoupling enumeration from selection.}
\ours relies on external RNG to uniformly sample ideas from LM-enumerated candidates, forming a sequence of randomized idea-level decisions. We show that this enumerate-select setup for idea sampling is critical. We compare \ours against a (1) no-enumeration variant and a (2) no-RNG variant. 

In the \textit{no-enumeration} variant, instead of instructing the LM to enumerate a list of plausible choices for each decision, we directly prompt the model for a single choice conditioned on the task and previous decisions. This ablation preserves the autoregressive decision structure, but variation in each choice now comes from the model's token distribution. One can also view this ablation as an autoregressive extension of Intent-Factored Generation.

In the \textit{no-RNG} variant, the LM still enumerates the same candidate options as in \ours, but we ask the model itself to choose randomly among them rather than using external RNG. Thus, the support of the conditional idea distribution is unchanged, but selection is now delegated back to the LM itself. As we saw earlier in \cref{tab:ablations} and again in \cref{tab:more_ablations}, removing external RNG sharply drops diversity. 
 
\paragraph{(C) Autoregressive idea sampling.}
\ours samples ideas autoregressively, where each decision is sampled conditioned on the previous ones. This allows earlier random choices to inform and diversify the options for later ones, enabling response diversity to come from the model's ability to enumerate ideas under different contexts as opposed to token entropy. We ablate \ours's autoregressive structure by comparing against two non-autoregressive variants. First, we enumerate $k$ complete ideas in a single context and sample one idea directly. This ``\textit{idea-list}'' variant can be viewed as enumerating directly from the joint distribution over ideas, and relies on token entropy to diversify across idea lists. Second, we decompose the task into the same semantic decision axes as \ours but sample each axis independently rather than sequentially. This ``\textit{independent-axes}'' variant approximates the joint distribution as a product of marginals rather than a chain of conditionals, and does not enable one decision to diversify the options for others. Results in \cref{tab:ablations,tab:more_ablations} show that \ours significantly outperforms both variants. 
\section{Details on \ours}
\label{appdx:method}
We present the prompt templates for each step in the Gacha harness, followed by concrete domain-dependent template values we use in our main experiments. \texttt{\{task\}} is always the user prompt
$x$. Model replies are parsed as JSON. Each LM call is performed at the default sampling parameters of the model (typically temperature 1.0, with reasoning on, see \cref{appdx:exp_setup:models}) as documented in the corresponding model cards.

\paragraph{Step 1: Plan the decision axes.}
To sample a response $y$ to some prompt $x$, we begin by planning the decision axes with the prompt in~\cref{prompt:planner}. We keep the first $T$ decision axes from the plan, where $T$ is a hyperparameter (see below for concrete per-domain values). We did not extensively sweep $T$ for each domain, and believe that optimizing this hyperparameter may lead to additional gains in future work.

\begin{promptbox}[label={prompt:planner}]{Planning decision axes}
\begin{Verbatim}[breaklines=true,breaksymbolleft={},fontsize=\small]
Design a schema that partitions the space of possible outputs for the task below into sequential decision points, so that distinct paths through the schema yield {goal}. Use AT MOST {T} questions total. Each decision point is phrased as a question, one per turn; its options are enumerated later, conditioned on all earlier answers.

Rules:
- Only partition what the task leaves free; never re-decide what it fixes.
- First brainstorm the set of questions, and then commit to the ordering. A good ordering should place questions with the fewest dependencies first, and leave the largest room for variation in later questions.
- Prefer simple, enumerable questions at the start, deferring finer-grained, abstract, or interpretive questions for later.
- Each question must introduce new high-level information that meaningfully steers the output.
- Give each question a two-or-three-word name for labeling its answer.

Task: {task}
Output exactly: {"questions": [{"name": "...", "ask": "..."}]}
\end{Verbatim}
\end{promptbox}

\begin{itemize}[leftmargin=2em,itemsep=1pt,topsep=2pt]
  \item \textbf{Story:} $T=4$; \texttt{\{goal\}} = ``stories that vary widely
  in: genre, setting, narrative tone, narrator and point of view, premise,
  emotional resolution, dominant imagery, main character''. These attributes are chosen to align with the evaluation attributes (see~\cref{sec:mainexps}) and to inform the LM of how the diversity of its stories will be judged. We include these attributes for all baseline prompting approaches as well to ensure fair comparison.
  \item \textbf{Chat:} $T=4$; \texttt{\{goal\}} = ``diverse high-quality
  responses''.
  \item \textbf{Image:} $T=6$; \texttt{\{goal\}} = ``diverse, high-quality
  generated images that would appear perceptually distinct even at a glance.''
  For image we use a condensed version of the planner that drops the
  \textbf{Rules} block, as we found a simpler prompt worked well in this
  setting; we view further per-domain prompt optimization as future work.
  
\end{itemize}

\paragraph{(Autoregressive Loop) Steps 2--4: Enumerate options for decision $t$, sample one choice, update the context for future decisions.} Now that we have planned the decision axes, we make the decisions autoregressively within a multi-turn LM conversation (separate context from the earlier planning stage). To pick a choice $z_t$ for the decision axis at time $t$, we prompt the model with the turn in~\cref{prompt:decision_turn}, appended to a growing multi-turn conversation. For the first decision at time $t=1$, since there is no previous decision, we prepend the original prompt $x$.

\begin{promptbox}[label={prompt:decision_turn}]{Decision turn $t$}
\begin{Verbatim}[breaklines=true,breaksymbolleft={},fontsize=\small]
Chosen {decision_name_(t-1)}: z_(t-1)

{decision_question_t} List at least {m} distinct options. Make sure that each option is consistent with all previously made choices.
Output exactly:
{"options": ["...", "...", ...]}
\end{Verbatim}
\end{promptbox}

At each step, the model returns a candidate set of enumerated options $\mathcal{C}_t$ as a JSON, from which the harness draws a single option $z_t$ uniformly with an RNG tool. This decision is then used to construct the prompt for the next turn deciding $z_{t+1}$. By adopting a multi-turn setup, we ensure that subsequent decisions are steered by all earlier decisions $z_{\leq t}$, enumerated options, and sampled outcomes.
If the LM call errors on any single turn (e.g., if the returned JSON is unparseable), then that decision is simply skipped.

For \textbf{story} and \textbf{image}, we use lists of size $m=32$. For \textbf{chat}, we use $m=24$. Again, this is not thoroughly optimized; other choices of this hyperparameter may yield stronger results.

\paragraph{Step 5: Realize.}
One final menu turn (\cref{prompt:idea_turn}) is appended to the same multi-turn conversation, over complete ideas rather than a single axis, and sampled the same way ($k=8$ in all domains). A fresh call then renders the prompt and all sampled choices into the response (\cref{prompt:executor}).

\begin{promptbox}[label={prompt:idea_turn}]{Idea turn (final turn within multi-turn conversation)}
\begin{Verbatim}[breaklines=true,breaksymbolleft={},fontsize=\small]
Chosen {decision_name_T}: {z_T}.

Given every choice above, list {k} distinct creative angles, themes, or approaches for the task. Be specific and varied.

Task: {task}
Output exactly:
{"options": ["...", "...", ...]}
\end{Verbatim}
\end{promptbox}

\begin{promptbox}[label={prompt:executor}]{Executor (fresh context)}
\begin{Verbatim}[breaklines=true,breaksymbolleft={},fontsize=\small]
Your task: {task}

Planned decisions:
{decision_name_1}: {z_1}
...
{decision_name_T}: {z_T}
Overall approach (the guiding take — write the whole response as this): {z_idea}

Write the single best {artifact} that satisfies the request while (a) satisfying the task and all its constraints and (b) realizing every design choice above. If a choice cannot be honored within the request's exact constraints, sacrifice the choice, not the request. Give the answer itself without preamble or meta-commentary.
\end{Verbatim}
\end{promptbox}

Here, \texttt{\{artifact\}} is replaced with ``story'' / ``response'' / ``caption'' for the story / chat / image domains, respectively.
\section{Details on Experimental Setup}
\label{appdx:exp_setup}

\subsection{Baselines and Prior Approaches}
\label{appdx:exp_setup:baselines}
We detail our baselines, including exact prompts used to implement them. In general, we place method-specific instructions into the system prompt (following the original works) and the user turn is simply the evaluation task prompt $x$. As with \ours, unless otherwise noted, every LM call uses the model's default sampling parameters.

\paragraph{Naive IID direct sampling.} The bare task is sent as the user turn with no additional system prompt (creative writing tasks still receive the shared story guidance as the system turn). We draw $n$ independent generations.

\paragraph{List prompting.} We prompt the model to generate $k$ complete responses as a numbered list. To get $n$ samples, we perform $\lceil n/k \rceil$ total independent calls and then pool the parsed responses. We use $k=8$ for creative writing and $k=32$ for the chat and image domains. We arrived at these choices of $k$ by extensive hyperparameter optimization, and find that it is critical for the list baseline's performance. Specifically, in initial experiments, we swept $k \in \{8,16,32,64,128,256\}$ up to $n=1024$ across many open-weight models and generally found that these specific choices of $k$ worked best across models. Intuitively, the reason the creative writing domain uses a much lower value of $k$ is because it demands much longer output---the LM cannot reason over many hundred-word stories at once. The system prompt is shown in~\cref{prompt:list}.

\begin{promptbox}[label={prompt:list}]{List baseline (system turn)}
\begin{Verbatim}[breaklines=true,breaksymbolleft={},fontsize=\small]
You are creative and can generate diverse, distinct outputs.

Generate {k} diverse responses to the task given by the user. Number them 1-{k}. Each should be meaningfully distinct from the others, so that the set of responses covers the full range of high-quality possibilities.
\end{Verbatim}
\end{promptbox}

\paragraph{Verbalized Sampling.} We follow the implementation of the original work~\citep{zhang2025verbalized}. A single call asks the model to verbalize a distribution of $k$ responses and their corresponding model-estimated probabilities. We make $\lceil n/k \rceil$ independent calls to cover $n$ samples. The probabilities are discarded after parsing. For all models except Inkling, we use $k=8$ for story and $k=32$ for chat and image, which we optimize in the same manner as our list baseline above. On Inkling, we use $k=8$ across the board, as we find the model underproduces responses at larger $k$. For creative writing, we additionally include a response length constraint in the prompt with a target of 400 words, following the original work's creative writing setup. We find that without this constraint, the generated stories are single-sentence snippets as opposed to full stories. For the chat and image domains, this length constraint is omitted. The system prompt is shown in~\cref{prompt:verbalized}.

\begin{promptbox}[label={prompt:verbalized}]{Verbalized Sampling (system turn)}
\begin{Verbatim}[breaklines=true,breaksymbolleft={},fontsize=\small]
Generate {k} responses to the input prompt. [Each response should be approximately {target_words} words.]
Return the responses in JSON format with the key: "responses" (list of dicts). Each dictionary must include:
- 'text': the response string only (no explanation or extra text).
- 'probability': the estimated probability from 0.0 to 1.0 of this response given the input prompt (relative to the full distribution).
Randomly sample the responses from the full distribution.
\end{Verbatim}
\end{promptbox}

\paragraph{Direct Planning (Intent-Factored Generation)}~\citep{ahmed2025intent,yao2019plan}. We generate each response in two stages, implemented as two separate LM calls. First, we independently sample a plan or intent for the response at a higher-than-typical temperature (1.2), using the brainstorming prompt in~\cref{prompt:ifg_brainstorm}. Then, in a second LM call, we execute the plan into a full response at the default temperature (\cref{prompt:ifg_execute}). With this method, diversity over responses comes from diversity over plans, which in turn comes from token sampling.

\begin{promptbox}[label={prompt:ifg_brainstorm}]{Intent-Factored Generation stage 1: generate plan (system turn; temperature 1.2)}
\begin{Verbatim}[breaklines=true,breaksymbolleft={},fontsize=\small]
Brainstorm keywords or ideas for responding to the task given by the user: the core content, angle, and approach a response could take. Give only a short list of keywords or ideas — not the response itself.
\end{Verbatim}
\end{promptbox}

\begin{promptbox}[label={prompt:ifg_execute}]{Intent-Factored Generation stage 2: execute (system turn, then user turn)}
\begin{Verbatim}[breaklines=true,breaksymbolleft={},fontsize=\small]
[system]
Respond to the task given by the user, guided by the provided keywords and ideas. Write the full response itself. Give the core answer only — no preamble, no explanation, no meta-commentary, no labels or headers. Strictly respect any length constraints in the task.

[user]
Task: {task}

Keywords and ideas: {plan}
\end{Verbatim}
\end{promptbox}

\paragraph{Persona prompting}~\citep{ge2024scaling, santurkar2023whose}. We generate each response by instructing the LM to respond as if it were a specific persona (see~\cref{prompt:persona}), where the persona is uniformly sampled (without replacement, to ensure independent responses to the same prompt use different personas) from the \texttt{persona} column of the Nemotron-Personas-USA dataset~\citep{nvidia/Nemotron-Personas-USA}.

\begin{promptbox}[label={prompt:persona}]{Persona (system turn)}
\begin{Verbatim}[breaklines=true,breaksymbolleft={},fontsize=\small]
Persona:
{persona}

Produce one response to the task as it would be produced by the persona above.

Use the persona to guide substantive choices: what possibilities are noticed, what approach is taken, which details matter, what tradeoffs are accepted, and how the response is structured.

Rules:

- Satisfy every explicit requirement in the task.
- Only use the persona to decide aspects that the task leaves open.
- The persona is the author or decision-maker, not necessarily a character or subject in the response.
- Do not mention, describe, or explain the persona.
- Do not merely alter vocabulary, tone, or surface style.
- Commit to the response this persona would most naturally produce, rather than trying to be generically comprehensive or broadly appealing.
- Produce only the final response.
\end{Verbatim}
\end{promptbox}

\paragraph{Seed Prompting (String Seed of Thought)}~\citep{misaki2026string}. We use the original work's prompt verbatim (see~\cref{prompt:ssot}), in its seed-injection variant. For each sample, we generate a fresh 24-character random ASCII string. The specific vocabulary we draw over is all upper and lower-case letters, digits 0-9, and punctuation from \verb|!@#$%^&*()-_=+|.

\begin{promptbox}[label={prompt:ssot}]{Seed prompting (system turn)}
\begin{Verbatim}[breaklines=true,breaksymbolleft={},fontsize=\small]
You are a helpful AI Assistant designed to provide well-reasoned and detailed responses. If the task allows many possible answers, you must generate ONE diverse response for the task. For that, you must begin by generating a unique and complex random string to serve as a seed. This random string should appear sufficiently complex and unpredictable, with no obvious structure or pattern. Use your judgment to ensure it looks arbitrary and unguessable. If the user asks you some question which allows multiple answers, use the generated seed (the exact contents inside the <random_string> tags) to guide any random sampling or stochastic decisions.

Follow these steps for every instruction:
1. Output the random seed string enclosed within <random_string> and </random_string> tags.
2. Think deeply and carefully about the user's question, and enclose this reasoning within tags. You have to generate ONE response leveraging the generated seed---the exact contents inside the <random_string> tags, to ensure your single answer is unique and diverse. Make sure to extract maximum randomness from the string by using all of its content.
3. Provide your final answer, enclosed within <answer> and </answer> tags.
Strictly follow this tag structure, and respond in the following format:
<random_string> ... </random_string> <answer> ... </answer>

A pre-generated random seed has been provided for you: {seed}
Do not generate your own random string. Use this exact seed as the contents of the <random_string> tags, then follow the steps above to derive your answer from it.
\end{Verbatim}
\end{promptbox}

\paragraph{Shared prompt suffix for creative writing.} To ensure that generation aligns with evaluation, every method (including \ours, see~\cref{appdx:method}) is told which attributes the story diversity evaluation measures. For baselines on our creative writing domain, we append the text in~\cref{prompt:story_guidance} to each baseline's system prompt.

\begin{promptbox}[label={prompt:story_guidance}]{Story guidance (appended to the system turn for creative writing tasks)}
\begin{Verbatim}[breaklines=true,breaksymbolleft={},fontsize=\small]
When generating diverse stories, give particular weight to varying them along these axes: genre, setting, narrative tone, narrator and point of view, premise, emotional resolution, dominant imagery, main character. Do not override choices already fixed by the task.
\end{Verbatim}
\end{promptbox}

\subsection{Evaluated Models}
\label{appdx:exp_setup:models}
We serve all open-weight models on one H200 node with SGLang~\citep{zheng2024sglang}. We present the models evaluated and sampling parameters used in evaluation in \cref{tab:appdx_models}. In general, we use the default parameters recommended by the corresponding model card. We serve Inkling models at their native NVFP4, GPT-OSS models at their native MXFP4, and the remaining models at BF16.

\begin{table}[tbp]
\centering
\small
\setlength{\tabcolsep}{5pt}
\begin{NiceTabular}{@{}llll@{}}
\toprule
\textbf{Model} & \textbf{Sampling} & \textbf{Reasoning} & \textbf{Max tok.} \\
\midrule
Qwen 3.5 family~\citep{qwen3.5}         & top-$p{=}0.95$, top-$k{=}20$, presence penalty 1.5 & on      & 32{,}768 \\
GPT-OSS family~\citep{agarwal2025gpt}   & top-$p{=}1.0$                                  & medium  & 65{,}536 \\
Inkling family~\citep{inkling}          & top-$p{=}1.0$                                  & high    & 65{,}536 \\
Gemma 4 family~\citep{team2026gemma}    & top-$p{=}0.95$, top-$k{=}64$                   & on      & 32{,}768 \\
\texttt{gpt-5.5}~\citep{gpt55}          & API default                                    & default & 32{,}768 \\
\texttt{claude-opus-4-8}~\citep{opus48} & API default                                    & default & 32{,}768 \\
\bottomrule
\end{NiceTabular}
\caption{Models used in our experiments with their sampling parameters. Unless otherwise noted for specific experiments, all models are sampled at $T{=}1.0$. For open-weight models we adopt each model card's recommended generation settings with reasoning enabled; for API models we use the provider defaults (Opus 4.8's default reasoning is adaptive, high).}
\label{tab:appdx_models}
\end{table}

\subsection{Evaluation}
\label{appdx:exp_setup:evaluation}
We present details on evaluations for each domain, including any LM judge prompts used. We use Qwen3.6-27B as our LM judge with thinking on and its model-card recommended sampling parameters (same as the parameters for Qwen 3.5 models, see~\cref{appdx:exp_setup:models}). We use Qwen-3-Embedding-8B for text embeddings~\citep{zhang2025qwen3}.

\paragraph{In-the-wild chat.}
We evaluate on the test split of InfinityChat~\citep{jiang2026artificial}. We judge response validity with the prompt shown in~\cref{prompt:chat_validity}.

\begin{promptbox}[label={prompt:chat_validity}]{Chat validity judge}
\begin{Verbatim}[breaklines=true,breaksymbolleft={},fontsize=\small]
You are grading whether a RESPONSE is a valid and complete answer to a PROMPT.

The standard is a reasonable user's satisfaction: the RESPONSE is VALID 
if a user who issued the PROMPT would be happy accepting it as an answer -- it addresses everything they asked, is correct and coherent, and honors every constraint they stated.

Guidelines:

1. Explicit constraints are binding. If the PROMPT states a requirement
   (a count, format, topic, source, length, etc.), the RESPONSE must
   honor it exactly. "Give me exactly three slogans" means exactly three; "a detective created by Agatha Christie" rules out Sherlock Holmes.

2. Unstated preferences are not constraints. If the PROMPT asks for a
   slogan and the RESPONSE offers several good ones, that is VALID -- the user got what they asked for. Do not penalize length, style, tone, extra options, or extra explanation the user did not forbid.

3. Real things must be real. If the PROMPT asks to name or identify
   something that exists (in the real world or in real, published
   fiction), the RESPONSE must give a genuine instance; one invented by
   the responder is INVALID.

4. It must be a genuine, usable answer. A RESPONSE that is incoherent,
   off-topic, mostly placeholder, too incomplete to use, or that
   deflects the question back to the user without answering it is
   INVALID.

Respond with ONLY a JSON object: {"reason": "<one sentence>", "valid": true|false}.
Prompt: {prompt}
Response: {response}
\end{Verbatim}
\end{promptbox}

\paragraph{Creative writing.} We collect in-the-wild creative writing prompts by annotating and filtering Wildchat~\citep{zhao2024wildchat} with an LM judge to keep only writing requests. From this subset, we perform a manual quality pass to remove ill-formed or illegible prompts, and uniformly sample 50 requests as our final evaluation set. Five of these evaluation prompts are shown verbatim (typos preserved) in~\cref{prompt:creative_writing_examples}.

To measure diversity, we annotate each story along 8 semantic dimensions with the LM judge system prompt in~\cref{prompt:story_annotator}; the generated story itself is supplied as the user turn. The annotations are then embedded per-dimension (i.e., each story yields 8 dimensions $\rightarrow$ 8 embeddings), and we measure final diversity as average Vendi score across the dimensions, following~\citet{springer2026annotations}. To score quality, we directly use the EQ-Bench Creative Writing judge prompt~\citep{paech2023eq}, following prior work~\citep{zhang2025verbalized}. We reproduce this prompt in~\cref{prompt:story_quality} for reference.

\begin{promptbox}[label={prompt:story_annotator}]{Creative writing annotation judge (system turn)}
\begin{Verbatim}[breaklines=true,breaksymbolleft={},fontsize=\small]
You are a careful literary analyst. You read a short story and describe it along several dimensions. For each dimension, give a SHORT DESCRIPTION--a few words to a single sentence--that captures what is actually distinctive about THIS story on that dimension. Do not use a fixed vocabulary or force the story into broad buckets: if two stories differ subtly on a dimension, your descriptions should differ subtly too. Base every description only on the text given; do not invent details that are not there. If a dimension is genuinely absent or indeterminate, say so briefly (e.g. "none" / "unclear"). Before describing, FIRST judge whether the text is actually a complete short story. Set "valid_story" to false if it is instead a one- or two-sentence description/synopsis of a story, a bare outline or list of beats, a refusal, empty/garbled output, or otherwise not a real written-out narrative; set it to true only for an actual story.

The dimensions:
- genre: the story's genre or mode.
- setting: where and when it takes place (time, place, world).
- narrative_tone: the overall mood/voice.
- narrator: the narrative perspective and who tells it.
- premise: the core situation or what the story is fundamentally about, in one phrase or sentence.
- resolution_emotion: the emotional note the ending lands on.
- dominant_imagery: the most prominent recurring images, motifs, or sensory texture.
- main_character: who the central character is--a brief descriptor.

Respond with ONLY a JSON object. Its first key is "valid_story" (boolean). Its remaining keys are exactly: genre, setting, narrative_tone, narrator, premise, resolution_emotion, dominant_imagery, main_character — each value is your description for that dimension. No extra keys, no commentary outside the JSON.
\end{Verbatim}
\end{promptbox}

\begin{promptbox}[label={prompt:story_quality}]{Creative writing quality judge (EQ-Bench rubric, from \citet{paech2023eq})}
\begin{Verbatim}[breaklines=true,breaksymbolleft={},fontsize=\tiny]
You are an expert in assessing creative writing. Your task is to score the quality of the test model's response below, by several metrics, on a 0-20 scale.

[PROMPT START]

{writing_prompt}

[PROMPT END]

[TEST MODEL RESPONSE]

{test_model_response}

[TEST MODEL RESPONSE END]

[Task]

You are an expert in assessing creative writing. Your task is to score the quality of the test model's response below, by several metrics, on a 0-20 scale.

Scoring notes:

- A scores of 20 represents a masterpiece.

- If the criteria is not relevant to the piece (e.g. criteria relating to dialogue but the piece has none), exclude it from your assessment.

- Do not penalize NSFW output; that is the default for some models. Just assess it on its merits.

- Everything within the "TEST MODEL RESPONSE" section was written by the test model. Sometimes models like to write comments on the piece after the piece is concluded; if this happens you should ignore their comments.

- You are to write a comprehensive analysis of the piece, then give your scores.

- If the output is not a story but rather a one or two sentence description or premise of a story, you should score it very low on all criteria, as it is not a complete story.

- You are a critic, and your job is to be critical, especially of any failings or amateurish elements.

- After the analysis, output the scores as a SINGLE JSON object mapping each criterion's key to its integer 0-20 score. Output nothing after the JSON object.

- Output format is:

[Analysis]

Write your detailed analysis here.

[Scores]

{
  "surprising_creative": <0-20>,
  "imagery_descriptive": <0-20>,
  "nuanced_characters": <0-20>,
  "emotionally_complex": <0-20>,
  "elegant_prose": <0-20>,
  "earned_light_dark": <0-20>,
  "emotionally_engaging": <0-20>,
  "consistent_voice_tone": <0-20>,
  "natural_flow": <0-20>,
  "overall_engagement": <0-20>
}

---

Now, rate the supplied model output on the following criteria (use the JSON key shown for each):

Surprising and Creative  (key: "surprising_creative")
Imagery and Descriptive Quality  (key: "imagery_descriptive")
Nuanced Characters  (key: "nuanced_characters")
Emotionally Complex  (key: "emotionally_complex")
Elegant Prose  (key: "elegant_prose")
Well-earned Lightness or Darkness  (key: "earned_light_dark")
Emotionally Engaging  (key: "emotionally_engaging")
Consistent Voice/Tone of Writing  (key: "consistent_voice_tone")
Sentences Flow Naturally  (key: "natural_flow")
Overall Reader Engagement  (key: "overall_engagement")
\end{Verbatim}
\end{promptbox}

\begin{promptbox}[label={prompt:creative_writing_examples}]{Creative writing task examples (5 of 50, from WildChat)}
\begin{Verbatim}[breaklines=true,breaksymbolleft={},fontsize=\small]
1. Tell me a story about three cats who went to Australia on a spring break trip. Please make the story funny and full of valuable life lessons.
2. Write a short story story about two neighbors
3. Write a story with moon idioms.
4. Write a story of a girl who likes to run
5. Write a story about someone who has no magic in a world of magic but is so physically fit that he is stronger than many of those who can use magic
\end{Verbatim}
\end{promptbox}

\paragraph{Image generation.} We filter OneIGBench for open-ended queries via an LM judge. Many of the prompts in OneIGBench are highly specific, and admit only a few valid images. We filter out these close-ended prompts in three stages. First, we exclude the text-rendering and knowledge-reasoning prompt categories, where the image content is essentially fully specified (e.g., \texttt{Generate the first page of the paper from the following information: Title: Denoising Diffusion Probabilistic Models}...). Next, we use an LM judge to annotate each prompt as open- or close-ended and remove all close-ended prompts. Finally, we pass all remaining prompts to the LM judge in a single context, and ask it to return the 50 most open-ended prompts. Five randomly sampled prompts from the resulting set are shown verbatim in~\cref{prompt:image_gen_examples}.

\begin{promptbox}[label={prompt:image_gen_examples}]{Image generation task examples (5 of 50, from OneIG-Bench)}
\begin{Verbatim}[breaklines=true,breaksymbolleft={},fontsize=\small]
1. A young child, seated at a lively restaurant table, eagerly observes the vibrant, fragrant dishes of genuine, traditional cuisine being served, captured in a lifelike style that showcases real people.
2. Swiss buildings, tall, deep underground view.
3. Aquarium expansion extends oceanward at sunset in futuristic long beach.
4. On a rooftop overlooking a sprawling city, two casually dressed individuals are seated. One is drinking from a cup, the other on phone.
5. Two characters hold hands; one dressed in a dark outfit with short black hair, the other adorned in elaborate white and red armor. The image unfolds in an artistic blend reminiscent of anime-style storytelling.
\end{Verbatim}
\end{promptbox}

When rendering images, we use Qwen-Image-2512~\citep{wu2025qwenimagetechnicalreport} with 30 denoising steps, true-CFG scale 4.0, and at 1328$\times$1328 resolution. We prompt the image model with ``\texttt{\{Original caption\}\textbackslash nExtra details:$\,\,$\{LM-rewrite\}}'' as the positive prompt, and the default model-recommended negative prompt (which essentially just steers the model away from low-resolution outputs). We embed each image with DreamSim for diversity evaluation, using the best-performing ensemble model from the original work~\citep{fu2023dreamsim}.
\section{Details on protein design pilot study}
\label{appdx:protein}
\subsection{Task overview}
We compare \ours with other inference-time methods for diversifying LM generations in a de novo protein binder design workflow. We task the LM to generate a set of diverse captions describing generic candidate binders with different folds and binding-interface configurations, and use a fixed text-conditioned protein generation model~\citep{lee2026prompteina} to realize each caption in complex with a specified downstream target. We evaluate the resulting designs by measuring predicted interface confidence after refolding each binder sequence jointly with its target, and structural diversity among designs that pass an interface confidence threshold.

\begin{table}[htbp]
\centering
\small
\setlength{\tabcolsep}{6pt}
\begin{NiceTabular}{llc}
\toprule
\textbf{Target} & \textbf{Hotspot residues} & \textbf{Binder length} \\
\midrule
BHRF1  & A65, A74, A77, A82, A85, A93 & 80--120 \\
IFNAR2 & A45, A73, A75, A77, A89, A91 & 60--175 \\
PD1    & A33, A95, A97, A102          & 80--150 \\
TrkA   & X294, X296, X333             & 50--120 \\
PDL1   & A37, A39, A49, A98           & 64--155 \\
DerF21 & A10                          & 70--185 \\
IL7RA  & B58, B80, B139               & 50--120 \\
VEGFA  & W81, W83, W91                & 50--140 \\
\bottomrule
\end{NiceTabular}
\caption{Targets considered in our protein binder design pilot~\citep{pacesa2025one}. Hotspots are specified by chain ID and residue number in the input structure. Binder lengths are sampled uniformly from the ranges shown (inclusive).}
\label{tab:appdx_binder_targets}
\end{table}

We specifically consider eight benchmark targets following \citet{pacesa2025one}: BHRF1, IFNAR2, PD1, TrkA, PDL1, DerF21, IL7RA, and VEGFA. For each target, we use the sequence and structure, predefined hotspot residues specifying the intended binding site, and binder-length range supplied by the benchmark configuration (documented in \cref{tab:appdx_binder_targets} for reference). We sample binder lengths uniformly from the target-specific range; these target-specific inputs are then supplied as conditioning to the protein generation model alongside the generated captions. Note that we reuse the same length draws across caption generation methods to ensure fair comparison. Also note that these target-specific inputs are not specified to the caption-generating LM---initial results showed that across methods, omitting this information led to better performance. All methods receive the same shared captioning task prompt, shown in \cref{prompt:protein_task}, which includes few-shot examples.

\begin{promptbox}[label={prompt:protein_task}]{Protein binder design task}
\begin{Verbatim}[breaklines=true,breaksymbolleft={},fontsize=\small]
Generate a caption describing a designed protein binder: a single-chain protein that folds on its own and binds a target protein at a defined epitope on that target's surface. The binder's CATH class must be either mainly alpha or alpha beta. Describe the binder itself -- its fold and the interface it forms with the target. Do not name a target protein, do not specify which epitope is bound, and do not state the binder's length; those are fixed by the design task.

Each caption should be factual and descriptive, using concrete numbers and specific structural terms. Write each caption as a bulleted list, one attribute per line, in the form '- <attribute>: <value>', using the attribute names from the category list verbatim. No speculative language, no prose paragraphs, no commentary.

Example captions for reference style (match this format exactly):
1.
- CATH class: mainly alpha
- CATH architecture: up-down bundle
- helix count: 3
- sheet count: 0
- compactness and shape: compact globular
- surface hydrophobicity level: low outside the interface
- interface residue count: 24
- interface buried surface area: 890 square Angstrom
- interface hydrophobic contact count: 17
- interface H-bond count: 4
- interface geometry type: flat helical face
- interface shape complementarity score: 0.71

2.
- CATH class: alpha beta
- CATH architecture: two-layer sandwich
- helix count: 2
- strand count: 5
- sheet count: 1
- beta topology type: antiparallel meander
- contact order value: 18
- disulfide bond count: 1
- disulfide topology type: intra-chain
- interface residue count: 31
- interface buried surface area: 1140 square Angstrom
- interface H-bond count: 11
- interface salt bridge count: 3
- interface geometry type: concave beta face wrapping a protruding loop
- interface shape complementarity score: 0.66
\end{Verbatim}
\end{promptbox}

Using this prompt, we generate binder captions with GPT-5.5, one of the two frontier models among the models we considered. In preliminary experiments, Opus 4.8 often declined requests involving biological design. We compare \ours with direct sampling, list prompting, and verbalized sampling, whose method-specific system instructions are detailed in~\cref{appdx:exp_setup:baselines}. Direct sampling provides a naive reference, and list prompting and verbalized sampling were generally the best baselines on GPT 5.5 in our main experiments. We generate 1024 captions with each method. 

\paragraph{Method hyperparameters.} Following our main experiments, we use $k=32$ for list prompting and verbalized sampling. For \ours, we use 7 decision steps and prompt for 16 enumerated options per decision. The goal string used in \ours's schema prompt is ``protein structure captions that vary widely.''

\subsection{Evaluation of binder captions} Given each generated caption, we pair it with a binder length sampled from the target-specific binder-length range, the target's sequence and structure, and a fixed set of hotspot residues defining the desired epitope, and pass these inputs to a fixed text-conditioned protein structure generation model~\citep{lee2026prompteina}. The target is represented as residue tokens carrying its sequence and structural features, while the hotspot mask enters through target sequence and pair features to identify residues the binder should contact. The caption is encoded through two complementary frozen language-model pathways: Qwen-3-Embedding-8B produces a pooled representation that is projected and broadcast to every binder residue through adaptive layer normalization, providing global conditioning, while Qwen-3-8B produces token-level representations that the protein states attend to in each denoiser layer, preserving finer-grained structural and interface constraints. Starting from Gaussian noise for the binder's $\mathrm{C}_\alpha$ coordinates and per-residue side-chain latents, the model keeps the target fixed and generates the binder by simulating its learned stochastic differential equation for 400 Euler--Maruyama steps with classifier-free guidance. A frozen variational autoencoder then decodes the sampled representation into an amino acid sequence and all-atom target--binder complex. We generate one complex per caption while holding the target, hotspots, binder-length range, and sampling configuration fixed across caption generation methods.

To evaluate predicted binding, we refold each generated binder sequence jointly with its target using ESMFold2-Fast~\citep{candido2026language} and compute ipSAE$_{\min}$ from the resulting predicted aligned error matrix~\citep{dunbrack2025res}. For each residue pair spanning the predicted binder--target interface, ESMFold2 estimates the error in their relative positions; ipSAE converts these errors into a score between 0 and 1, with low predicted error yielding a high score, and adjusts the normalization for interface size. ipSAE$_{\min}$ takes the smaller of the asymmetric binder-to-target and target-to-binder scores, so a high value requires confident inter-chain geometry in both directions and serves as a proxy for predicted binding strength rather than a direct affinity measurement. We operationally designate designs with ipSAE$_{\min}>0.5$ as successful. To measure diversity among successful designs, we remove the shared target chain and compare the generated binders within each target using all-vs-all Foldseek with exact TM-align~\citep{van2024fast,zhang2004scoring}. TM-score measures global structural similarity after optimal three-dimensional alignment, normalized for protein length: it ranges from 0 to 1, with larger values indicating more similar folds and 1 indicating identical structures. We symmetrize the pairwise TM-scores and use the resulting similarity matrix as the Vendi kernel. The Vendi score summarizes the spectrum of this matrix and can be interpreted as the effective number of structurally distinct binders, such that higher values indicate greater structural diversity. We compute diversity separately for each target so that differences between target structures or epitopes do not contribute to the score.

\end{document}